\documentclass[11pt]{article}

\usepackage[preprint]{acl}

\usepackage{enumitem}
\usepackage{float}
\usepackage{booktabs}
\usepackage{times}
\usepackage{latexsym}
\usepackage[most]{tcolorbox}
\usepackage{verbatim}
\usepackage{subcaption}
\usepackage[T1]{fontenc}

\usepackage[utf8]{inputenc}
\usepackage{amsfonts}
\usepackage{amsmath}
\usepackage{amssymb}
\usepackage{microtype}

\usepackage{dblfloatfix}
\usepackage{inconsolata}
\usepackage{caption}
\usepackage{graphicx}

\title{What's the Point? 
Spatial Grammar \&  Index Resolution\\for Sign Language Recognition}

\author{Oline Ranum$^1$, Simon Hadfield$^1$ and Richard Bowden$^1$ \\
  $^1$Centre for Vision, Speech and Signal Processing, University of Surrey, Guildford, UK \\
  \texttt{\{o.ranum, s.hadfield, r.bowden\}@surrey.ac.uk}} 

\begin{document}
\maketitle

\begin{abstract}
Sign language models are predominantly trained with gloss-sequence or text supervision, thereby under-modeling non-lexical and productive constructions. One comparatively tractable instance is spatial \emph{indexing}: pointing gestures that assign discourse entities to spatial loci for subsequent co-reference, which lexicon-centric objectives largely fail to capture. We present a targeted evaluation of indexing in Sign Language Recognition, showing that despite comprising 10-15\% of signing content, 
indexing is poorly recovered. We introduce a framework for training and evaluating indexing experts, establishing a baseline for index-aware sign language modeling. Our approach decomposes spatial reference resolution into index detection and discourse entity linking. The resulting mention representations enable automatic annotation and non-lexical structure modeling, and serve as an auxiliary indexing expert that augments a frozen SLR model at inference time.


\end{abstract}

\section{Introduction}
\label{sec:intro}
Most Sign Language Recognition (SLR) and Translation (SLT) pipelines map sign language to gloss-label sequences \cite{zhou2021backtranslation,zhou2021multicue,yin2021simulslt,chen2022twostream,zhang2023sltunet,bragg2019sign,DeCoster2024} or text \cite{zhou2023glossfree,wong2024sign2gpt,chen2024factorized,asasi2025beyond,chen2025c2rl,kim2025mllm,sincan2025glossfree}. This reflects supervision in current benchmarks, where signed phrases are aligned with lexical or coarse lexicon-like tokens, often under an assumption of one-to-one correspondence. Implicitly, this framework assumes that language in both modalities can be represented as a linearly ordered sequence drawn from a finite discrete lexicon \cite{brown2026,DeCoster2024}. This assumption captures only part of the linguistic structure of sign languages \cite{baker2016linguistics,bragg2019sign}.

In spontaneous signed discourse, approximately 40\% of signs are non-lexical, consisting of productive constructions that exploit three-dimensional space, iconicity, and context \cite{fenlon2014}. A particularly frequent and tractable instance of non-lexical sign classes is \emph{spatial indexing} (pointing), whereby signers establish entities and roles by associating them with spatial loci and subsequently refer back to them through pointing signs \cite{Liddell1980,Kegl1987,Cormier2010}. 

Although visually simple at the local level, indexing depends on discourse state and productive use of space: similar manual forms can serve pronominal, locative, or determiner-like functions, and referents may be reassigned as narratives evolve \cite{VanHoek1992,EmmoreyLilloMartin1995,Ozyurek2010,Cormier2013}. Systems trained primarily under lexical objectives are therefore not optimized to (i) reliably distinguish indexing from lexically similar signs and (ii) maintain referential consistency across phrases.

We curate a training and evaluation framework for index-awareness, using linguistic corpora and an established benchmark with manually annotated pointing. We evaluate a state-of-the-art SLR model, which performs poorly in recovering pointing-related glosses. This motivates an explicit \emph{indexing expert} that complements lexically oriented SLR models with targeted mechanisms for detecting indexing events and tracking discourse referents.

In this work, we present a modular pipeline for \textbf{index detection} and \textbf{discourse entity linking}. We decompose spatial reference into two stages: index detection and entity linking, separating local recognition from discourse-level tracking. Concretely, we implement (i) an Index Proposal Network (IPN) that identifies indexing segments, and (ii) a differentiable online Entity Linking Module (ELM) that incrementally clusters detected mentions into entity representations. The resulting state supports both corpus-scale annotation and downstream tasks. Our contributions are as follows.
\begin{enumerate}[noitemsep]
\item We show that spatial indexing is under-modeled by lexicon-based objectives.

\item We curate a training and evaluation framework for index-awareness in SLR.

\item We propose a two-stage index modeling approach using an Index Proposal Network and an online memory module for discourse-level entity tracking.

\item We introduce a targeted evaluation protocol for index-awareness and show that the resulting system improves indexing recovery and reduces WER as an auxiliary module for a frozen SLR model at inference time.
\end{enumerate}

\subsection{Task Definition}
\label{sec:units}

We model spatial reference in signed discourse at three levels:
(i) sign-level \emph{index detection},
(ii) document-level \emph{entity linking} over detected mentions, and
(iii) optional \emph{expert integration} into a downstream SLR model.

\paragraph{Input Representation.}
A sign language document is represented as an ordered sequence of $T$ temporally localized segments,
$
D = (x_1, \dots, x_T),
$
where each segment $x_t$ corresponds to a short clip aligned to a gloss-level annotation boundary and is represented by pose features extracted from the underlying video frames. The segment granularity remains fixed throughout; the processing mode differs across stages as described below.

\paragraph{Index Detection.}
For each segment $x_t$, the IPN predicts whether the segment realizes an indexing sign. Segments predicted as indexing are treated as index mentions and passed to the linking stage.

\paragraph{Entity Linking.}
Detected index mentions are processed sequentially in document order. The model maintains a memory of discourse entities and assigns each mention either to an existing entity or to a newly created one, inducing a clustering over mentions within the document. 

\paragraph{Index-Supported CSLR.}
At inference time, we incorporate index proposals and predicted entity assignments as an auxiliary bias for a frozen SLR model. Segments align with the evaluation units of the target gloss-based benchmark and are processed sequentially in document order. The index confidence score increases the relative weight of pointing-related tokens when indexing is detected, and the entity memory state promotes consistency across repeated mentions.

\section{Background and Related Works}
\subsection{Non-Lexical Signs and Discourse}
Sign languages consist of both established and productive lexicons. Lexical signs are relatively conventionalized forms with stable phonological parameters and meanings \cite{baker2016linguistics}. While language-dependent, they  account for approximately 60\% of signs in discourse \cite{fenlon2014}. The remaining content consists largely of non-lexical and productive constructions, including pointing signs, depicting constructions, and enactment \cite{brown2026}, which exploit the three-dimensional signing space around the signer \cite{baker2016linguistics, DeCoster2024}. Through spatial location, movement, iconicity, and embodiment, they encode reference, structure events, and represent participants or actions. Unlike fully lexicalized signs, their interpretation is more context-dependent and emerges from the interaction between articulatory form and discourse structure.

\subsection{Spatial Indexing and Annotations}
Indexing signs establish reference by assigning discourse entities to spatial loci in signing space \cite{Liddell1980,Kegl1987}, a mechanism observed across many sign languages \cite{Aronoff2003Classifier,MathurRathmann2010Agreement}. Beyond their pronominal use, where they refer to previously introduced entities \cite{VanHoek1992, EmmoreyLilloMartin1995, Cormier2010}, they also serve locative or determiner-like functions (e.g., “here”, “that”) \cite{Ozyurek2010, Cormier2013}. Interpretation depends on discourse context rather than local appearance. Referents may be grounded in present entities or assigned to abstract loci for absent ones, and once introduced, loci can be reused via pointing, gaze, or agreement \cite{EmmoreyLilloMartin1995,Schlenker2018}. Unlike spoken pronouns, indexing signs lack fixed grammatical features (e.g., gender, number); instead, meaning derives from spatial loci and evolving discourse, which may also be reassigned over time \cite{Cormier2010,Yin2021}. Despite their prevalence (10–16\% of tokens; Table~\ref{tab:datasetproperties}), indexing signs are inconsistently and coarsely annotated in SLR and SLT benchmarks, often collapsing functions into gloss labels and omitting referential structure. This limits discourse-level evaluation and motivates treating indexing as a structured prediction task: detecting indexing events and resolving their associated entities.

\subsection{Sign Language Recognition}

SLR maps signed input, typically RGB video or pose representations, to gloss labels \cite{bragg2019sign,DeCoster2024}. The task is commonly divided into isolated SLR (ISR), where each clip contains a single sign \cite{jiang2024signclip, wong2025signrep, ranum2024b, wu2025cross, hu2021signbert, zhao2024masa, jiang2021, ranum2024a}, and SLR, where unsegmented signing streams must be transcribed into ordered gloss sequences \cite{camgoz2018neural, guo2025bridging, raude2024, cihan2017subunets, hao2021self, KollerO20160919, pu2019iterative,cui2019deep}. Progress has been driven by stronger visual backbones and transformer-based sequence models, as well as retrieval-style objectives that align signs and text. However, available benchmarks largely rely on subtitle-aligned interpreted broadcasts or gloss-annotated datasets from controlled laboratory environments \cite{brown2026, DeCoster2024}, thereby encouraging lexical sequence modeling rather than explicit discourse-aware modeling. Spatially grounded phenomena such as indexing are therefore rarely treated as structured reference mechanisms and are typically collapsed into gloss tokens.

\subsection{Coreference Resolution and Memory-Based Entity Modeling}

Coreference resolution in NLP clusters textual mentions into entity representations \cite{lee2017endtoend,joshi2020spanbert}. Memory-augmented models maintain bounded, online-updated entity representations to handle long documents \cite{henaff2017referential,xia2020learning}. Similar to our approach, but designed for spoken language, \citet{sundar2024} decouples mention detection from entity assignment and updates a working memory via learned ranking, a decomposition we adapt to visually grounded discourse. \citet{Yin2021} explored sign-language coreference over gloss annotations; we instead operate on the visual modality and generate supervision automatically with an LLM. We use the term \textit{linking} rather than coreference resolution, as indexing encompasses not only anaphoric reference, but also locative and determiner-like functions realized through spatial loci.

\section{Method}
\label{sec:method}
\begin{figure*}[!h]
\centering
\includegraphics[width=0.75\linewidth]{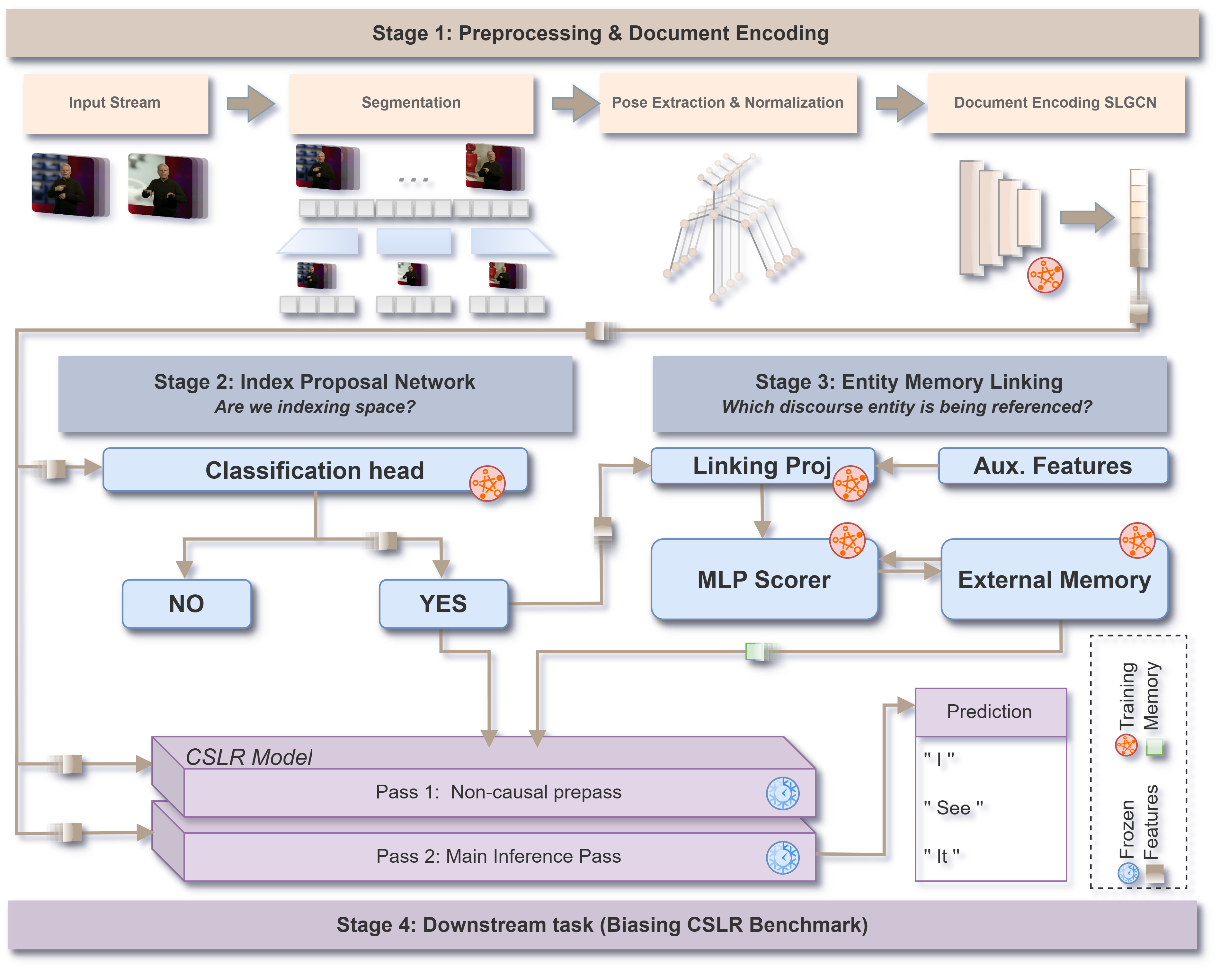}
\caption{Overview of the proposed pipeline. Gloss-level pose segments are encoded with the SLGCN, then processed by the IPN. Detected mentions are passed to the ELM module, which maintains an external entity memory. The resulting signals can be injected as inference-time biases into a frozen SLR model.}
\label{fig:architecture}
\end{figure*} 
Our methodology models indexing as a two-stage process: (i) index detection from pose-based segments, and (ii) discourse entity linking via an online memory mechanism. The resulting entity state can optionally be integrated to refine SLP models. Figure~\ref{fig:architecture} illustrates the pipeline.

\subsection{Pose Representation and Encoder}
\label{docenc}
Each segment is represented by poses extracted from video. Body pose is obtained with SMPL \cite{SMPL-X:2019} (8 upper-body 3D joints; Appendix~\ref{sec:feature_layout}) and hand pose via WiLoR \cite{potamias2025wilor} (21 joints per hand), forming a unified skeleton that captures handshape, motion, and spatial configuration. Poses are normalized for scale and viewpoint to support cross-dataset generalization. Features are extracted using the SL-GCN \cite{jiang2021}, which models frames as spatial graphs and captures temporal dynamics via spatio-temporal convolutions. The encoder outputs a segment embedding $h_g \in \mathbb{R}^d$ via global pooling, where $g$ indexes gloss segments and $d$ is the embedding dimension.

\subsection{Index Proposal Network (IPN)}

Given a segment embedding $h_g$, the IPN predicts whether the segment corresponds to an index sign. A lightweight classification head maps $h_g$ to an index confidence score $p_g \in [0,1]$. Segments exceeding a decision threshold $\tau$ are treated as index mentions and passed to the entity linking stage. The encoder and classification head are trained using gloss-level indexing annotations from linguistically curated corpora.
\subsection{Entity-Linking Memory (ELM)}

Detected index mentions are processed in gloss-segment order. We maintain a dynamically growing entity memory $\{e_k\}_{k=1}^{K}$, where $k$ indexes entities, building on memory-augmented coreference models~\cite{henaff2017referential,sundar2024}.

Each segment embedding $h_g$ is projected into a dedicated linking space to decouple entity resolution from index classification, yielding a mention embedding $m_g \in \mathbb{R}^{d'}$. Since the encoder is trained with a binary objective, its representations capture coarse pointing cues and may suppress fine-grained spatial information required to distinguish individual referents. We therefore augment $m_g$ with geometric features $f_g$ derived from pointing direction, spatial locus, and trajectory (Appendix~\ref{app:elm_features}).

Given the current memory, we compute compatibility scores with each entity $e_k$:
\begin{equation}
s_{g,k} = \mathrm{MLP}([m_g;\, e_k;\, m_g \odot e_k;\, f_g]),
\end{equation}
together with a fixed null score $s_{g,\varnothing}=0$ for introducing a new entity. The assignment is obtained via $\arg\max$ over $\{s_{g,k}\}_{k=1}^{K} \cup \{s_{g,\varnothing}\}$ and trained using a softmax cross-entropy objective with LLM-derived labels.

If a mention links to an existing entity $e_k$, we update its representation via gated interpolation:
\begin{equation}
\begin{aligned}
e_k &\leftarrow \alpha_g\, e_k + (1-\alpha_g)\, m_g, \\
\alpha_g &= \sigma\!\left(\mathrm{MLP}([e_k;\, m_g])\right),
\end{aligned}
\end{equation}
where $\alpha_g \in (0,1)$ depends on both the entity state and the incoming mention. Otherwise, $m_g$ initializes a new memory slot. During training, we use teacher forcing to mitigate error propagation.
\subsection{Inference-Time SLR Integration}
\label{sec:cslr_method}

We integrate IPN and ELM into a frozen SLR model, CSLR2~\cite{raude2024}. Each video is processed by CSLR2 to produce frame-level logits $\ell \in \mathbb{R}^{F \times V}$, where $F$ is the number of frames and $V$ the vocabulary size. Time-aligned gloss annotations partition each pose sequence into segments $g$. We introduce two inference-time interventions: (i) an IPN boost injecting index confidence into pointing-token logits, and (ii) an ELM prior biasing predictions via cluster-level entity assignments.

\paragraph{Detection boost $w_{\text{IPN}}$.}
Each gloss segment $g$ with temporal extent $[t_{\text{start}}^g, t_{\text{end}}^g]$ is uniformly sampled (12 frames) and processed by the IPN to obtain a confidence score $p_g \in [0,1]$. This score is broadcast to frames within the segment, forming a segment mask $\mathbf{q}^{(g)} \in \{0,1\}^F$. Let $\mathbf{v}_{\mathcal{P}}$ denote the indicator over pointing tokens. The logits are updated as:
\begin{equation}
\ell \leftarrow \ell + w_{\text{IPN}} \cdot \mathbf{q}^{(g)} p_g \mathbf{v}_{\mathcal{P}}^\top
\end{equation}

\paragraph{Entity Linking prior $w_{\text{ELM}}$.}
Segments with IPN score $p_g \geq \tau$ are assigned to ELM clusters. For each cluster $k$, we compute a consensus pointing token $v_k^\star \in \mathcal{P}$ using a non-causal full-episode prepass over CSLR2 predictions, aggregating votes across all segments in the cluster (including synonym-group expansion; Appendix~\ref{app:synonyms}). Let $\mathbf{e}_{v_k^\star} \in \{0,1\}^V$ denote its one-hot representation. Each segment $g$ in cluster $k$ receives:
\begin{equation}
\ell \leftarrow \ell + w_{\text{ELM}} \cdot \mathbf{q}^{(g)} \mathbf{e}_{v_k^\star}^\top
\end{equation}
\section{Experiments}
\label{sec:experiments}
\subsection{Datasets and Benchmark}
\label{sec:datasets}
\begin{table}[!t]
  \centering
  \small
  \setlength{\tabcolsep}{5pt}
  \begin{tabular}{lllll}
    \hline
    \textbf{Data} & \textbf{\# Doc / Sent} & \textbf{ V$_T$} / \textbf{V$_G$} & \textbf{\# T$_{G}$} & \textbf{\# T$_{Idx}$} \\
    \hline
    BSLCP  & 409 / 8K    & 5K / 7K   & 74.8K  & 11.4K \\
    BOBSL  &35 / 4.4K &35K / 5K & 32K  & 4.0K \\
    MDGS   &318 / 63.5K & 22K / 9K & 357K & 57K \\
    \hline 
  \end{tabular}

  \caption{Statistics from relevant corpora and benchmark. Alongside the number of documents and sentences, we report the vocabulary size of all text tokens V$_T$ and glosses V$_G$, and the number of gloss tokens T$_G$ and index-related tokens T$_{idx}$.}
  \label{tab:datasetproperties}
\end{table}

We use two annotated corpora (BSLCP~\cite{Schembri2014BSL}, MDGS~\cite{konrad2020meinedgs}) to train the modules. BSLCP mark indexing instances with a \texttt{"PT:"} prefix, and MDGS maps signs to coarse pronominal and non-specified indexing categories (\texttt{"Ich"}, \texttt{"Du"}, \texttt{"INDEX"}). 

For SLR evaluation, we use CSLR-TEST~\cite{raude2024}, a 6-hour subset of the subtitle-aligned SENT-TEST split of BOBSL~\cite{albanie2021bobsl} spanning 35 episodes. Annotators verified automatically segmented clips and assigned each a free-form English label with optional sign-type tags for non-lexical phenomena (e.g., \texttt{*D}: depictions, \texttt{*P}: pointing). The \texttt{*P} tag is broadly defined and may include non-referential signs with pointing handshapes (e.g., \emph{what}, \emph{why})~\cite{raude2024}. The CSLR2 baseline excludes most \texttt{*P}-tagged segments when evaluating, we retain them to analysing spatial indexing.

\paragraph{General indexing set (GIS).}
CSLR-TEST contains 4.7K \texttt{*P}-tagged segments. Of these, 3.2K indexing instances (approximately 10\% of all tokens) are not associated with any lexical gloss:
$
g_{\mathrm{GIS}} = \{ g \mid \mathrm{Type}(g) = \texttt{*P} \land \mathrm{Lex}(g) = \emptyset \}
$,
where Type($g$) is the non-lexical sign class and Lex($g$) is the lexical gloss label of segment $g$.

\paragraph{Indexing proxy subset (IPS).}
To enable entity-specific evaluation, we isolate \texttt{*P}-labelled segments that overlap temporally with referential glosses, yielding 856 instances. Since \texttt{*P} conflates referential and non-referential cases, we define a filtered subset using a curated index-candidate vocabulary spanning pronouns, reflexives, and deictic forms:
$
g_{\mathrm{IPS}} = \{ g \mid \mathrm{Type}(g) = \texttt{*P} \land \mathrm{Lex}(g) \in \mathcal{I} \},
$
where $\mathcal{I}$ = \{me, my, mine, you, your, he, her, our, they, their, myself, yourself, this, that, here, there, \textit{i, yours, him, his, she, it, its, we, us, ours, them, himself, herself, itself, themselves, these, those}\}. This defines the subset $g_{\mathrm{IPS}}$ with identifiable referents. Only the first 16 items in $\mathcal{I}$ are present in the benchmark vocabulary. Although most \texttt{*P} segments are excluded from CSLR2 evaluation, $g_{\mathrm{IPS}}$ is partially retained in the lexically oriented evaluation when corresponding tokens exist in the pretrained CSLR2 vocabulary.

\subsection{Automatic Entity Label Generation}
\label{sec:labelgen}

To obtain supervision for entity linking at scale, we generate cluster labels via a two-stage pipeline. Candidate mentions are extracted from transcripts using POS tagging and rule-based span selection, after which an LLM assigns entity cluster IDs at document level under structural constraints. The resulting clusters are refined by a rule-based post-processing step that merges fragments sharing unambiguous person-pointing labels per signer and document (Appendix~\ref{sec:label_gen}).

\begin{table*}[!b]
\centering
\setlength{\tabcolsep}{5pt}
\small
\begin{tabular}{lllllllll}
\hline
\textbf{Train} & \textbf{Eval} &
\textbf{BalAcc} & \textbf{Macro F1} &
\textbf{Prec$_{\text{Lex}}$} & \textbf{Rec$_{\text{Lex}}$} &
\textbf{Prec$_{\text{Index}}$} & \textbf{Rec$_{\text{Index}}$} \\
\hline

B & B
& $0.85$ & $0.85$
& $0.85_{\scriptstyle\pm0.01}$ & $0.86_{\scriptstyle\pm0.01}$
& $0.85$ & $0.85_{\scriptstyle\pm0.01}$ \\

B & M
& $0.76$ & $0.76$
& $0.83_{\scriptstyle\pm0.01}$ & $0.67$
& $0.72$ & $0.86_{\scriptstyle\pm0.01}$ \\

B & BOB
& $0.77_{\scriptstyle\pm0.01}$ & $0.68_{\scriptstyle\pm0.01}$
& $0.96$ & $0.87_{\scriptstyle\pm0.01}$
& $0.35_{\scriptstyle\pm0.02}$ & $0.66_{\scriptstyle\pm0.02}$ \\

\hline

M & B
& $0.75_{\scriptstyle\pm0.01}$ & $0.74_{\scriptstyle\pm0.01}$
& $0.69_{\scriptstyle\pm0.01}$ & $0.92_{\scriptstyle\pm0.01}$
& $0.87_{\scriptstyle\pm0.01}$ & $0.58_{\scriptstyle\pm0.02}$ \\

M & M
& $0.87$ & $0.87$
& $0.88$ & $0.86$
& $0.87$ & $0.88$ \\

M & BOB
& $0.72$ & $0.70_{\scriptstyle\pm0.01}$
& $0.95$ & $0.92_{\scriptstyle\pm0.02}$
& $0.41_{\scriptstyle\pm0.04}$ & $0.52_{\scriptstyle\pm0.02}$ \\

\hline

BM-cw4 & B
& $0.82$ & $0.81$
& $0.76$ & $0.93$
& $0.91_{\scriptstyle\pm0.01}$ & $0.70_{\scriptstyle\pm0.01}$ \\

BM-cw4 & M
& $0.85$ & $0.85$
& $0.81$ & $0.91$
& $0.90$ & $0.79$ \\

BM-cw4 & BOB
& $0.73_{\scriptstyle\pm0.01}$ & $0.72_{\scriptstyle\pm0.01}$
& $0.95$ & $0.94$
& $0.47_{\scriptstyle\pm0.01}$ & $0.52_{\scriptstyle\pm0.01}$ \\

\hline

BM & B
& $0.85$ & $0.85$
& $0.83_{\scriptstyle\pm0.01}$ & $0.89_{\scriptstyle\pm0.01}$
& $0.88_{\scriptstyle\pm0.01}$ & $0.82_{\scriptstyle\pm0.01}$ \\

BM & M
& $0.87$ & $0.87$
& $0.87$ & $0.86_{\scriptstyle\pm0.01}$
& $0.86_{\scriptstyle\pm0.01}$ & $0.88$ \\

BM & BOB
& $0.78_{\scriptstyle\pm0.01}$ & $0.69_{\scriptstyle\pm0.01}$
& $0.96$ & $0.87_{\scriptstyle\pm0.02}$
& $0.37_{\scriptstyle\pm0.03}$ & $0.68_{\scriptstyle\pm0.04}$ \\

\hline
\end{tabular}

\caption{IPN classifier performance (mean $\pm$ std over 3 seeds). B = BSLCP, M = MDGS, and BOB = BOBSL. $L$ and $I$ denote lexical and indexing classes, respectively. Decision threshold is set to 0.5; zero standard deviations are omitted for readability.}
\label{tab:phase1_metrics}
\end{table*}

\subsection{Baseline: Index Recovery in CSLR}
\label{sec:cslr_baseline}

As described in Sec.~\ref{sec:datasets}, the CSLR2 evaluation protocol excludes most non-lexical signs, including instances from $g_{\mathrm{GIS}}$. To analyse indexing behavior, we augment the evaluation sequences by reinserting these segments and re-evaluate using indexing-specific metrics. We consider two conditions: \textbf{BL$_{\text{Lex}}$}, which follows the official CSLR2 protocol (WER including $g_{\mathrm{IPS}}$), and \textbf{BL$_{\text{IR}}$}, which reinserts the excluded $g_{\mathrm{GIS}}$ segments. We report WER$_{\text{All}}$, WER$_{\text{Index}}$, and WER$_{\text{Lex}}$ as defined in Appendix~\ref{sec:wer_metrics}. For $g_{\mathrm{GIS}}$, correctness is defined as a prediction $l_p \in \mathcal{I}$, reflecting token-level index recovery rather than referential accuracy (Sec.~\ref{sec:error_analysis}).


Table~\ref{tab:prelim} shows that WER$_{\text{All}}$ increases from 69.5\% to 70.5\% after index restoration, while WER$_{\text{Index}}$ reaches 98.4 \%  and Index IoU drops to 6.1\%. These results indicate that indexing is strongly under-modelled by current SLR objectives and motivate an explicit modeling component. Our reproduced WER (69.5\%) differs from the 65.5\% reported in~\cite{raude2024} but matches the performance reported for the released model weights.

\begin{table}[!h]
  \centering
  \small
  \setlength{\tabcolsep}{5pt}

  \begin{tabular}{lllll}
    \hline
    \textbf{Exp.} & \textbf{IoU$_I$ (\%)}  &\textbf{WER$_{All}$} & \textbf{WER$_{Index}$} & \textbf{WER$_{Lex}$} \\
    \hline
    BL$_{Lex}$ & 15.8 & 69.5 & 91.3 & 77.5 \\
    BL$_{IR}$  & 6.1 & 70.5 & 98.4 & 77.5 \\
    \hline
  \end{tabular}
\caption{CSLR2 baseline and index-restored performance on BOBSL-CSLR test set. BL$_{\text{Lex}}$ follows the standard CSLR2 protocol, while BL$_{\text{IR}}$ reinserts excluded index segments. IoU$_I$ measures index segment overlap; WER$_{\text{All}}$, WER$_{\text{Lex}}$, and WER$_{\text{Index}}$ denote overall, lexical, and index-specific error rates.}
  \label{tab:prelim}
\end{table}

\subsection{Evaluation of Index Detection}
\label{sec:ipn_results}

We train the IPN on index-related labels from BSLCP and MDGS and evaluate within- and cross-dataset performance (Table~\ref{tab:phase1_metrics}). The model generalises well across datasets, suggesting that aspects of indexing signs transfer across sign languages despite signer and lexical variation. Joint training (BM) further improves robustness.

To reduce false positives that propagate into entity linking (\textit{lexical bleed}), we prioritise precision over recall. We introduce a class-weighted variant (cw4), upweighting non-index errors by a factor of 4 in the cross-entropy loss to penalise spurious index predictions. Combined with a high-confidence threshold ($\tau=0.90$; Appendix~\ref{sec:threshold}), this yields a more conservative model. The resulting BM-cw4 model is used for downstream SLR experiments.

\subsection{Evaluation of Entity Linking}
\label{sec:elm_results}

We train the ELM on automatically generated entity clusters (Sec.~\ref{sec:labelgen}) using indexed mentions from the frozen IPN. Table~\ref{tab:phase2_coref} reports clustering performance on a test set built from automatically labelled BSLCP data. All models use the BM~cw4 encoder, and we compare BSLCP-only training with joint training on BM.

Joint training on MDGS and BSLCP slightly improves Entity Cluster Accuracy (ECA $= 0.65$ vs.\ $0.62$) but yields similar F1 ($0.43$ vs.\ $0.44 \pm 0.01$), indicating better cluster structure but no gain in per-mention precision. This improvement does not transfer to downstream CSLR: BSLCP-only training achieves lower WER on BOBSL under the setup in the next section. We therefore use the BSLCP-trained linker for SLR integration due to better label alignment and reduced variability.

\begin{table}[!t]
\centering
\footnotesize
\setlength{\tabcolsep}{3pt}

\begin{tabular}{lccccc}
\hline
\textbf{IPN/ELM} & \textbf{ECA} & \textbf{F1} &
\textbf{WER$_{\text{All}}$} & \textbf{WER$_{\text{Lex}}$} & \textbf{WER$_{\text{Index}}$} \\
\hline

BM$_{cw4}$/B  & 0.62 & $0.44$ & 70.1 & 78.4 & 64.3 \\
BM$_{cw4}$/BM & 0.65 & $0.43$ & 70.2 & 78.4 & 64.7 \\
\hline
\end{tabular}

\caption{Entity linking and downstream SLR performance on BSLCP and BOBSL. BSLCP metrics are averaged over three seeds; BOBSL WER is reported using the checkpoint at $w_{\mathrm{IPN}}=10$, $w_{\mathrm{ELM}}=60$. ECA denotes entity cluster accuracy; WER$_{\text{All}}$, WER$_{\text{Lex}}$, and WER$_{\text{Index}}$ denote overall, lexical, and index-related word error rates, respectively. Standard deviations are reported where applicable.}
\label{tab:phase2_coref}
\end{table}
  
\begin{figure*}[!b]
  \centering
  \includegraphics[width=\linewidth]{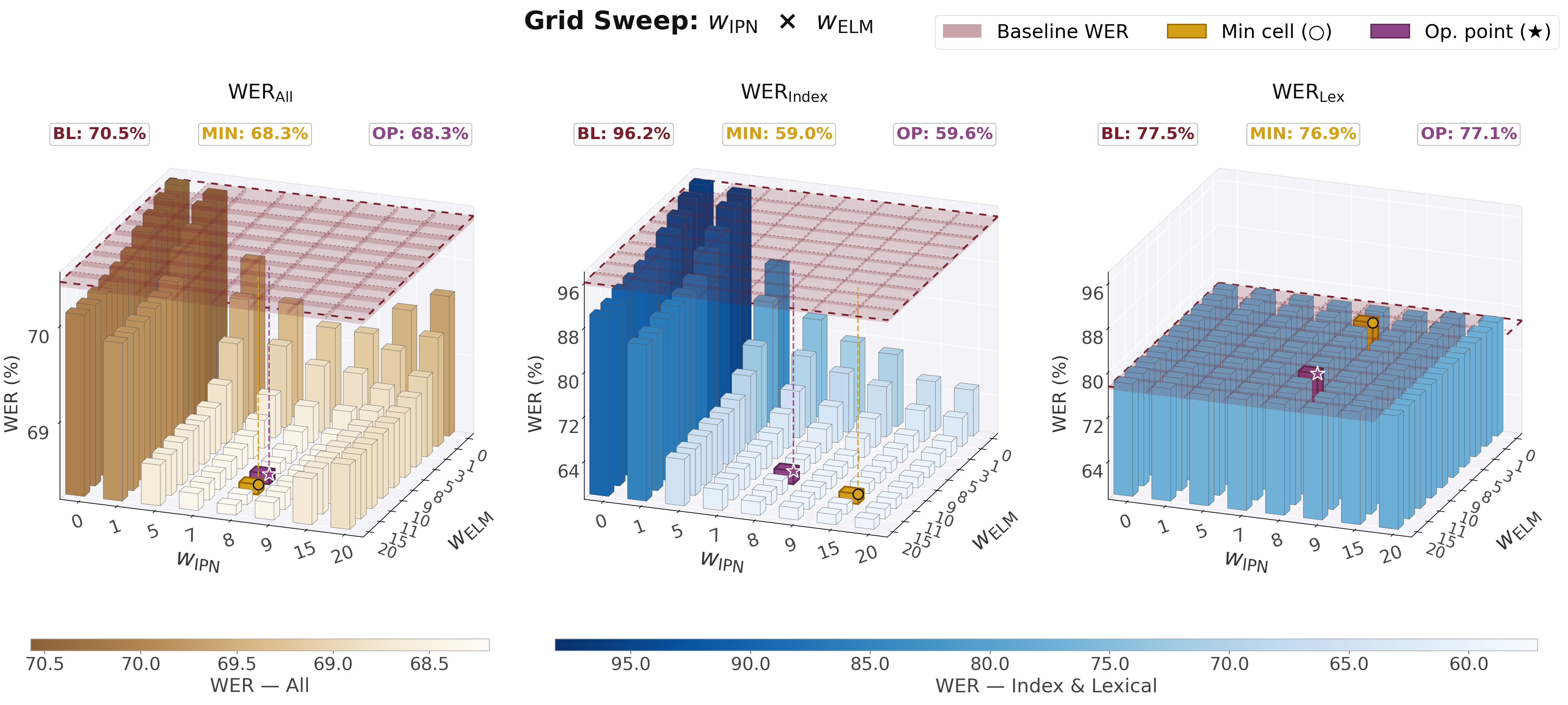}
\caption{Coreference and downstream SLR performance on BSLCP and BOBSL. BSLCP metrics are averaged over three seeds; BOBSL results use the checkpoint at $w_{\mathrm{IPN}}=10$, $w_{\mathrm{ELM}}=60$. Standard deviations are shown where applicable.}

  \label{fig:gridsearch}
\end{figure*}

\subsection{Index-Supported CSLR}
\label{sec:cslr_integration}

We integrate the IPN and ELM as inference-time biases into a frozen CSLR2 model. We evaluated using a standard word error rate (WER$_{\text{All}}$) and two diagnostic subgroup metrics isolating indexing behavior. Across all evaluations, out-of-vocabulary IPS glosses are mapped to in-vocabulary pointing-form equivalents (Appendix~\ref{app:synonyms}).

\paragraph{Trade-off between index recovery and lexical bleed.}
Figure~\ref{fig:gridsearch} shows a grid search over the detection boost $w_{\text{IPN}}$ and entity linking prior $w_{\text{ELM}}$. The heatmaps report (i) WER$_{All}$, (ii) WER$_{Index}$ (indexing sub-vocabulary), and (iii) WER$_{Lex}$ (lexical tokens). Increasing $w_{\text{IPN}}$ is the primary driver of WER$_{Index}$ reduction, while $w_{\text{ELM}}$ provides a further consistent gain, as shown in Table~\ref{tab:cslr_ablation}.However, large $w_{\text{IPN}}$ causes \emph{lexical bleed}, as misclassified non-index segments are biased toward pointing tokens, raising WER$_{Lex}$, creating a trade-off between index recovery and lexical accuracy that produces a visible valley in WER$_{All}$ around $w_{\text{IPN}}\approx 8$.

\begin{table}[t]
\centering
\small
\setlength{\tabcolsep}{4pt}
\begin{tabular}{llll}
\hline
\textbf{Ablation} & \textbf{WER$_{\text{All}}$} & \textbf{WER$_{\text{Index}}$} & \textbf{WER$_{\text{Lex}}$} \\
\hline
Baseline      & 70.5 & 96.3 & 77.4 \\
No-memory (IPN)     & 69.1 & 66.5 & 76.9 \\
No-proposal (ELM)  & 70.1 & 89.5 & 77.4 \\
IPN + ELM          & 68.3 & 59.6 & 77.1 \\
\hline
\end{tabular}
\caption{Ablation of IPN and ELM components in CSLR2. “No-memory (IPN)” removes entity linking memory while retaining index proposal detection, “No-proposal (ELM)” removes index proposal while retaining entity linking, and “IPN + ELM” corresponds to the full system with $w_{\text{IPN}}=8$ and $w_{\text{ELM}}=10$.}
\label{tab:cslr_ablation}
\end{table}

We select the operating point $w_{\text{IPN}}=8$, $w_{\text{ELM}}=10$ at the floor of this valley. Compared to the baseline (WER$_{Index}=96.2$, WER$_{All}=70.5$, WER$_{Lex}=77.5$), this configuration reduces WER$_{Index}$ to $59.6$ and WER$_{All}$ to $68.3$. WER$_{Lex}$ is slightly improved to $77.1$ due to the removal of false-positive lexical segments, with these gains outweighing the introduction of new false-positive index predictions, confirming that substantial improvements in indexing performance are achieved without degrading lexical performance.

\begin{figure*}[t!]
  \centering
  
  \begin{subfigure}{\linewidth}
    \centering
    \includegraphics[width=\linewidth]{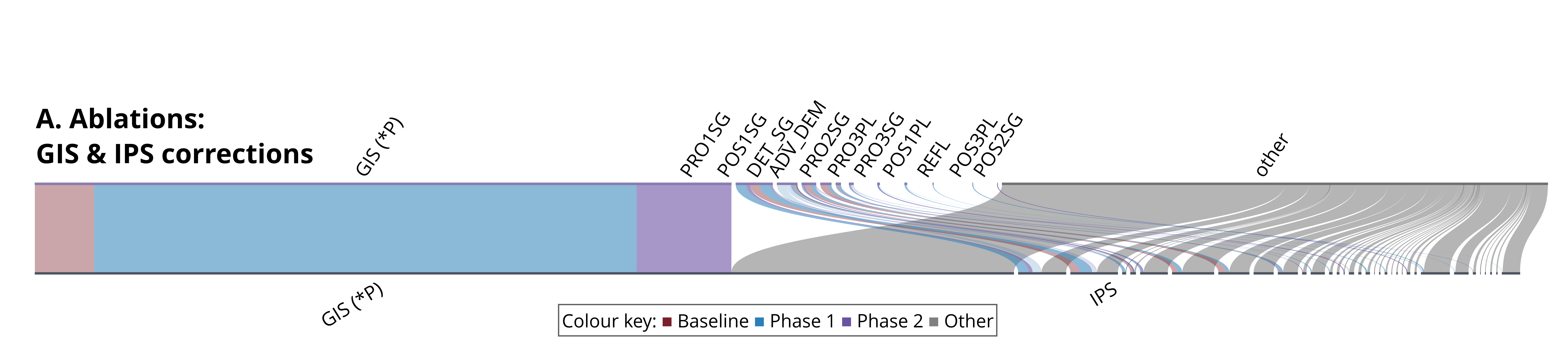}
  \end{subfigure}
  
  \vspace{-0.2em}
  
  \begin{subfigure}{\linewidth}
    \centering
    \includegraphics[width=\linewidth]{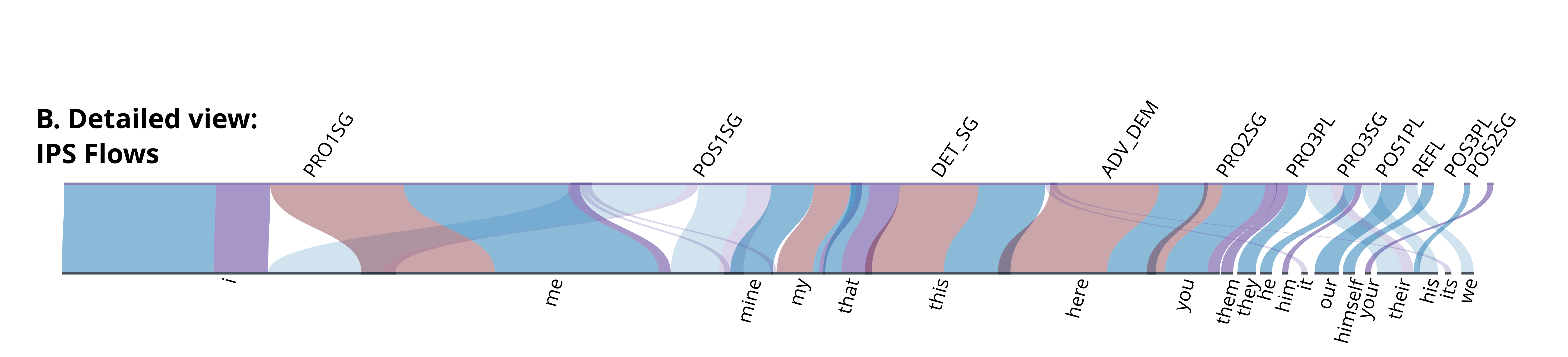}
  \end{subfigure}
\caption{Flow-based analysis of pointing-token predictions under the full system configuration ($w_{\text{IPN}}=8$, $w_{\text{ELM}}=10$). Each flow traces a ground-truth pointing token to the coarse referential class of the predicted token, grouped into pronoun and deictic categories (Appendix~\ref{app:synonyms}). The top panel shows all $g_{GIS}$ and $g_{IPS}$ instances, while the bottom panel isolates $g_{IPS}$ tokens to highlight entity-specific behavior. Flows terminating in the correct category indicate successful recovery of referential structure; remaining flows correspond to lexical predictions outside the pointing vocabulary. Compared to the baseline, a larger fraction of flows terminate in the correct referential classes, indicating that the external indexing expert improves recovery of pointing expressions.}
  \label{fig:sankey}
\end{figure*}

\begin{figure*}[!t]
    \centering
    \includegraphics[width=\linewidth]{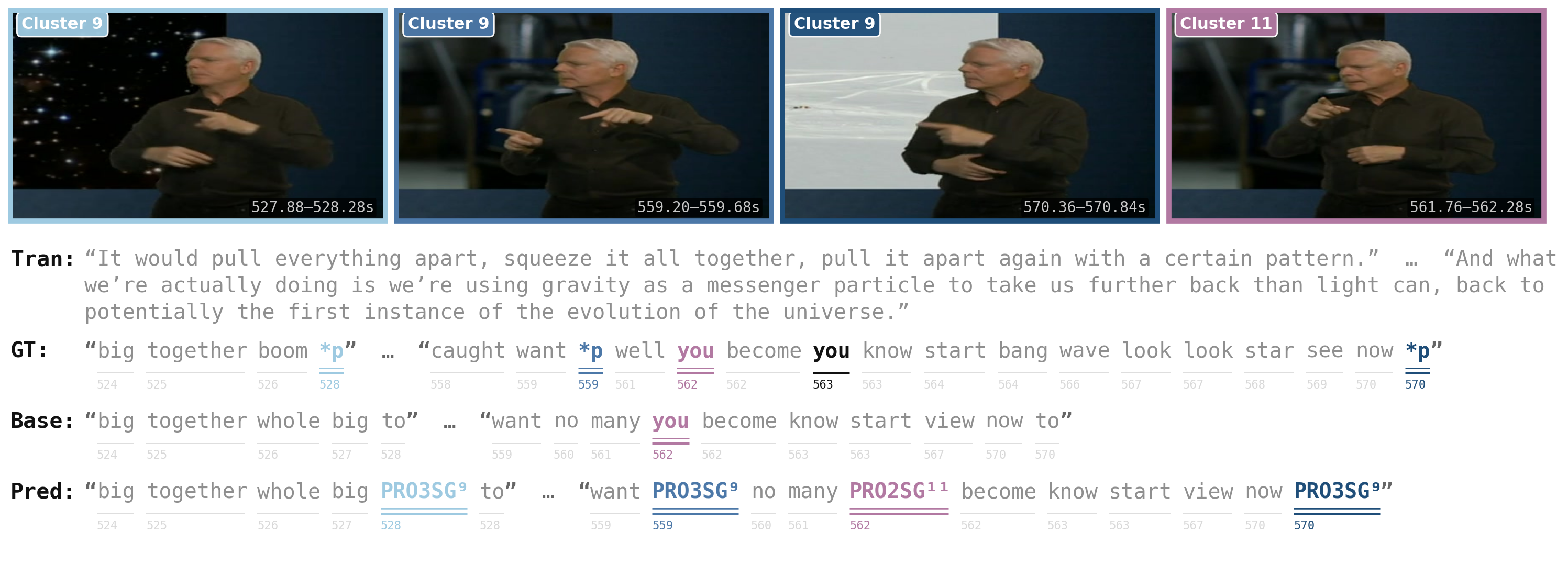}
    \caption{Samples from episode 6164207930460576679. The sample showcases both cross-sentence cluster recognition and in-sentence cluster separation between a third person singular entity and a second person 'you' being referenced. The base CSLR2 only picked up the 'you' index sign, but was not able to pick up any of the third-person references. The second reference to the second person is not detected. }
    \label{fig:qualitative}
\end{figure*}

Fig.~\ref{fig:sankey} presents a Sankey flow diagram for the full system ($w_{\text{IPN}}=8$, $w_{\text{ELM}}=10$). Each flow traces a ground-truth pointing token to the coarse referential category of the predicted token, grouped into pronoun and deictic classes (Appendix~\ref{app:synonyms}). The overview panel shows all GIS and IPS flows, while the detail panel isolates IPS tokens to highlight category distinctions. Under the baseline, only 6.1\% of ground-truth index tokens are recovered; the IPN detection boost alone raises this to 61.5\%, and the full system reaches 71.2\%, confirming that entity linking provides a consistent gain on top of detection. 

Figure~\ref{fig:qualitative} shows an example of a qualitative evaluation for the ELM, with additional results in Appendix \ref{sec:qual_phase2}. The model resolves clusters across sentences and captures subtle pointing distinctions, including second- and third-person forms. Failure cases include missed mentions and occasional over-clustering. Overall, the model identifies substantially more identity-related references than the baseline. We observe strong grouping of visually similar signs, with clusters often remaining consistent across sentences. However, errors include over-merging of spatially proximate entities and under-merging across long temporal ranges.

\section{Conclusion}

We presented a modular framework for spatial indexing in signed discourse, addressing a structural gap in lexicon-centric sign language modeling pipelines. The framework establishes index-aware modeling as a structured task over sign streams, rather than a by-product of SLR supervision.

We first showed that a strong CSLR2 model fails to recover indexing tokens despite competitive overall performance, motivating explicit modeling of index structure. Our approach decomposes the problem into index detection and discourse entity linking: a pose-based IPN generalises across data distributions and can be calibrated toward high precision, while a differentiable online ELM tracks discourse entities via geometric and learned features. Automatic cluster supervision from an LLM pipeline reduces the need for large-scale coreference annotations. When integrated as inference-time biases into a frozen CSLR2 backbone, the system reduces WER$_{\text{Index}}$ from 96.2 to 59.6 while leaving WER$_{\text{Lex}}$ stable, demonstrating that discourse-aware components can complement existing recognition systems without retraining. Future work will extend this treatment of spatial grammar to other productive constructions such as depicting signs and role shift, improve discourse-level supervision, for both recognition and translation models.

\clearpage

\section{Limitations}
\label{sec:error_analysis}

\paragraph{Distribution Shift Across Signing Contexts.} Training on conversational corpora from Deaf signers and evaluating on interpreted broadcast introduces substantial domain shift in factors including signer style and identity, recording conditions, and linguistic scope. While the training data consist of conversational signing by Deaf participants in controlled settings, BOBSL contains translated content produced by professional (often hearing) interpreters. This can lead to shifts in patterns of language use and references. For example, referents such as \textit{I} and \textit{you} may change dynamically over time, and pointing direction may reflect the source broadcast rather than the signer. Interpretation further introduces variability due to rapid decision-making under temporal constraints, which can alter referential structure and spatial grounding. 

\paragraph{Labels and evaluation metrics.} BOBSL annotations rely on automatic temporal segmentation with human verification, so ground-truth labels may be incomplete. For example, we observe some apparent false positives correspond to valid indexing signs absent from the annotations. This limits the reliability of quantitative evaluation and motivates complementary qualitative analysis.

In the BOBSL evaluation, the primary recovery metric for $g_{\mathrm{GIS}}$ measures the proportion of ground-truth \texttt{*P} segments for which any token in $\mathcal{I}$ is predicted. This definition is necessary because \texttt{*P} annotations do not encode explicit referential categories, preventing exact matching at the token level. As a consequence, semantically incorrect substitutions (e.g., predicting \textit{they} for a self-reference \textit{me}) are counted as correct under the $g_{GIS}$ evaluation. Therefore, WER$_{\text{Index}}$ captures index token presence rather than true referential accuracy and should be interpreted as a lower bound on semantic error. More precise evaluation would require referential annotations aligned with individual \texttt{*P} instances, which are currently only available for $g_{IPS}$.


\paragraph{IPN architectural limitations.} Detection errors primarily arise from lexical signs with index-like handshape configurations or transient pointing-like movements (e.g.\ \texttt{"what"}, \texttt{"one"}, \texttt{"why"} in BSL). Missed detections are caused by occlusion, coarticulation, and imprecise temporal boundaries, and can propagate into broken entity-linking chains. A common failure mode is confusion between personal and possessive forms (\textit{I} vs.\ \textit{my}): frontal occlusion obscures index-finger extension, while singular–plural distinctions, encoded in subtle 3D trajectory variations, are confounded by coarticulation and are not reliably captured. Cross-linguistic variation (e.g.\ between DGS and BSL) in the handshape realisation of possessive forms introduces additional challenges for multilingual training. Several informative cues not yet integrated include gaze direction and torso shifts; viewpoint normalisation may further suppress the latter.

\paragraph{ELM architectural limitations.}
Cluster supervision derived from the LLM pipeline is generated without a visual signal, so coreference chains that are visually grounded but linguistically ambiguous introduce noise, particularly in longer documents. The $s_{t,\varnothing}$ parameter, which controls the threshold for opening new entity slots, impacts cluster granularity: more negative values produce fewer, broader clusters, while less negative values fragment the entity space. We explored both fixed $s_{t,\varnothing}$ offsets and a learned $s_{t,\varnothing}$ head, but the absence of ground-truth coreference chains on BOBSL makes rigorous calibration intractable. In the downstream SLR experiment, $s_{t,\varnothing}$ primarily regulates cluster grouping sensitivity; WER converges across $s_{t,\varnothing}$ settings at moderate cluster weight, and a principled calibration method remains an open problem.

\paragraph{Qualitative samples and evaluation.} Cluster formation is further constrained by the underlying recognition pipeline, and entity clusters are not assigned explicit contextual names. Automatic entity naming remains an open challenge, and is expected to improve as sign language recognition models become more robust.

SLR integration operates purely at inference time by biasing token logits, without joint training of the recognition backbone. While this enables plug-and-play deployment, it limits the ability to learn direct index-aware representations. Consequently, improvements in index recovery translate only modestly to WER$_{All}$, partly due to the class imbalance between indexing and lexical tokens.

\bibliography{custom}

\clearpage

\appendix

{\Large\bfseries Supplementary Material}

\vspace{1.5em}

\noindent\textbf{Appendix Overview}
\vspace{0.3em}

\begin{itemize}[leftmargin=*, itemsep=0.2em, label={}]
    \item \ref{app:conceptualill} Conceptual Illustration of Method
    \item \ref{app:error_analysis} Error Analysis: IPN Index Proposal
    \item \ref{sec:qual_phase2} Qualitative Results: ELM BOBSL
    \item \ref{app:sweeps} Numerical Values of SLR Grid Sweeps
    \item \ref{sec:label_gen} Automatic Entity Label Generation
    \item \ref{sec:feature_layout} Input Feature Layout and ELM Auxiliary Features
    \item \ref{sec:threshold} IPN Threshold Ablation
    \item \ref{sec:wer_metrics} WER$_{All}$, WER$_{Index}$ and WER$_{Lex}$
    \item \ref{app:synonyms} Synonym-Based Evaluation Mapping
    \item \ref{app:guardrails} SLR Integration: Inference-Time Guardrails
    \item \ref{app:hyperparameters} Hyperparameters
    \item \ref{app:sizebudget} Model Size and Budget
    \item \ref{app:disclaimer} Disclaimer: Use of AI-Assisted Development
    \item \ref{app:license} License and Code 
\end{itemize}

\section{Conceptual Illustration of Method}
\label{app:conceptualill}
\begin{figure}[!b]
  \centering
  \includegraphics[width=2\linewidth]{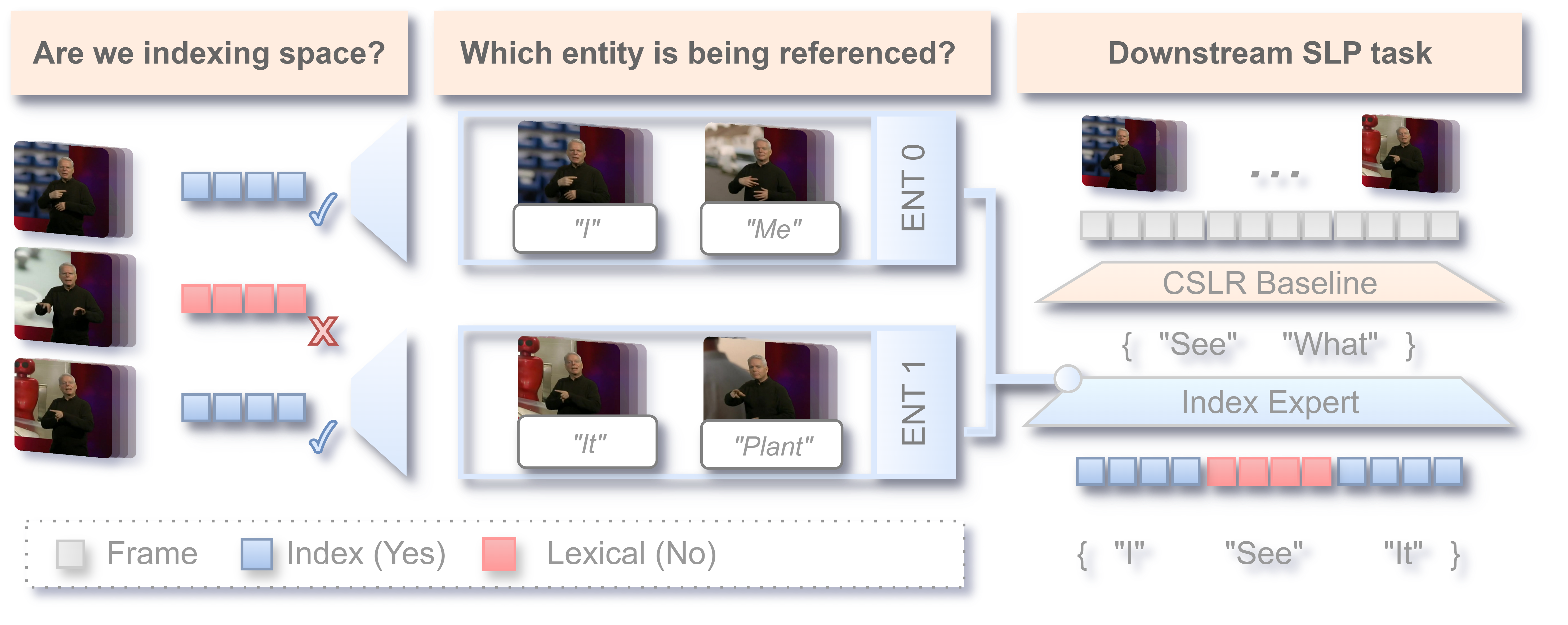}

  \vspace{0.5em}

  \makebox[\textwidth][c]{%
    \parbox{1.9\textwidth}{%
      \caption{Indexing is a grammatical function where discourse entities are associated with spatial\\loci in the signing space and subsequently re-referenced through pointing.}
      \label{fig:frontpage}
    }
  }
\end{figure}

Figure \ref{fig:frontpage} showcases a conceptual illustration of the proposed method. 

\section{Error Analysis: IPN}
\label{app:error_analysis}

\begin{figure*}[]
\centering
\setlength{\tabcolsep}{3pt}

\caption{Examples of false positives (two left columns) and false negatives (two right columns).}
\label{fig:phase1_fp_fnb}
\begin{tabular}{cccc}

{\tiny Q} & {\tiny R} & {\tiny S} & {\tiny T} \\
\includegraphics[width=0.24\textwidth]{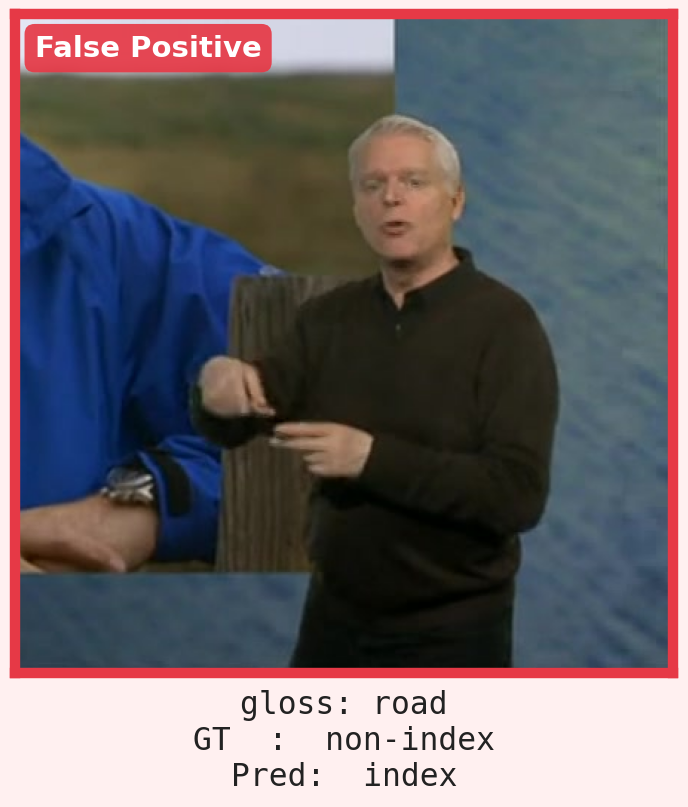} &
\includegraphics[width=0.24\textwidth]{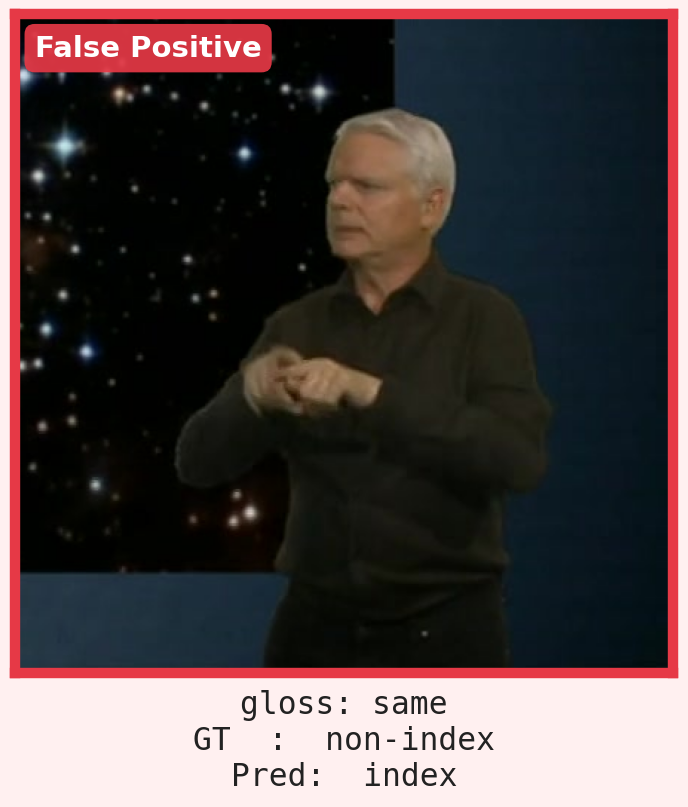} &
\includegraphics[width=0.24\textwidth]{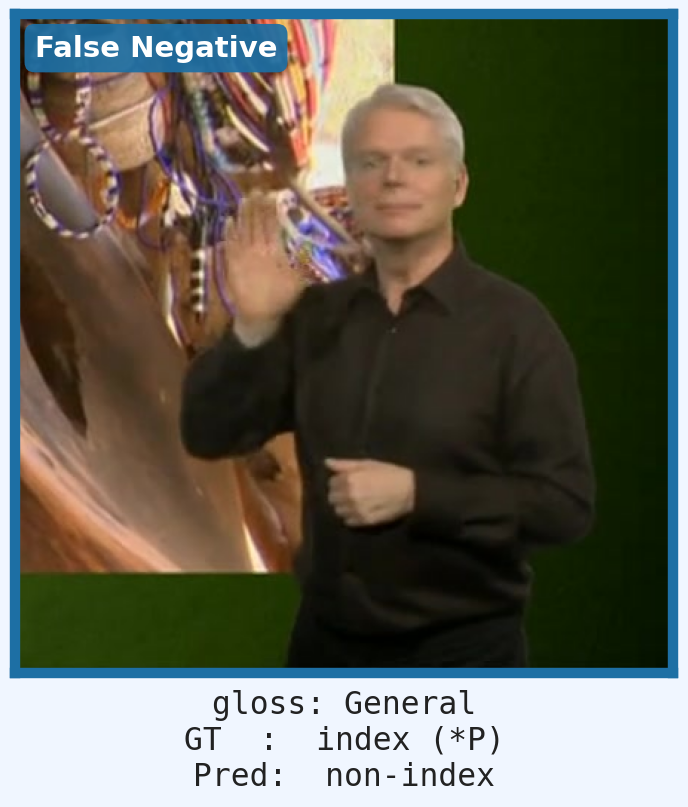} &
\includegraphics[width=0.24\textwidth]{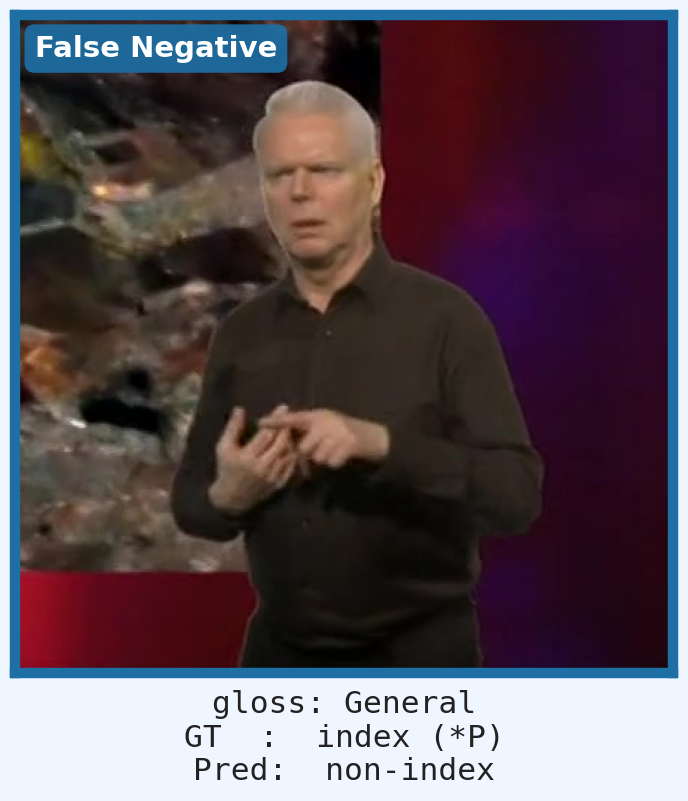} \\

{\tiny A} & {\tiny B} & {\tiny C} & {\tiny D} \\
\includegraphics[width=0.24\textwidth]{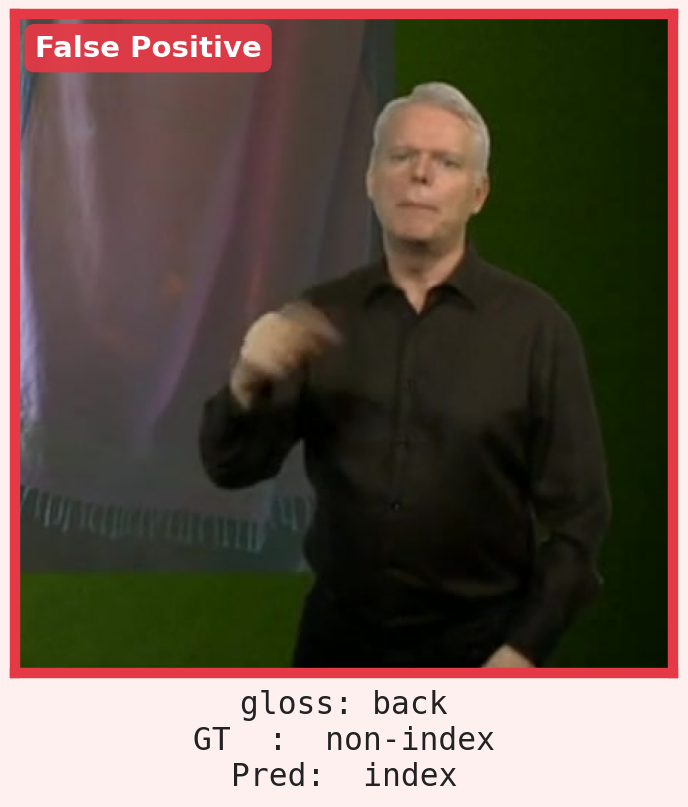} &
\includegraphics[width=0.24\textwidth]{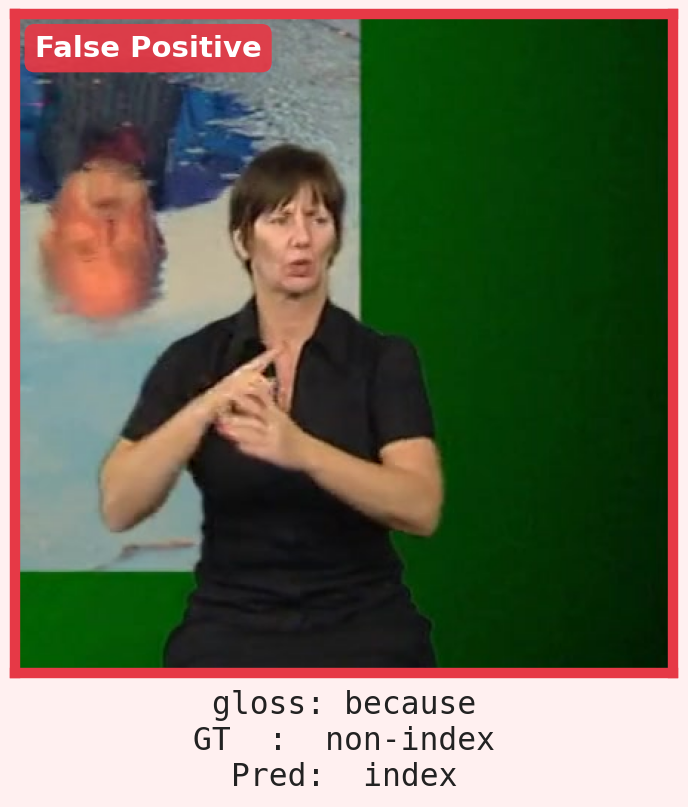} &
\includegraphics[width=0.24\textwidth]{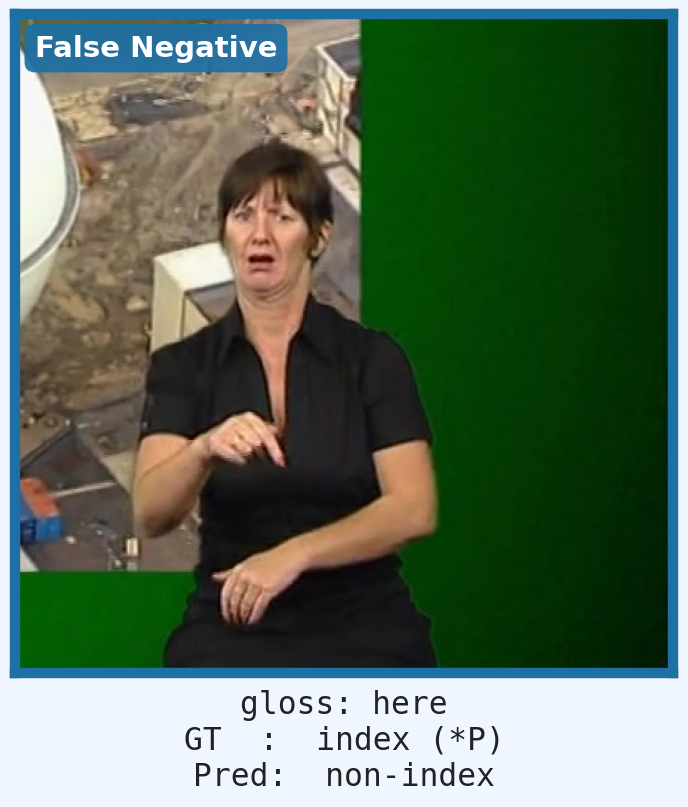} &
\includegraphics[width=0.24\textwidth]{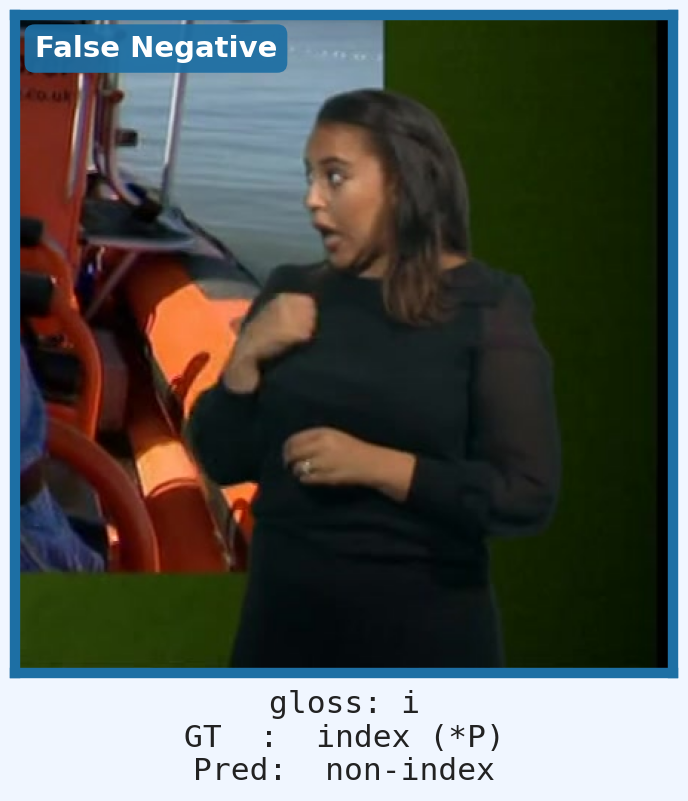} \\

{\tiny E} & {\tiny F} & {\tiny G} & {\tiny H} \\
\includegraphics[width=0.24\textwidth]{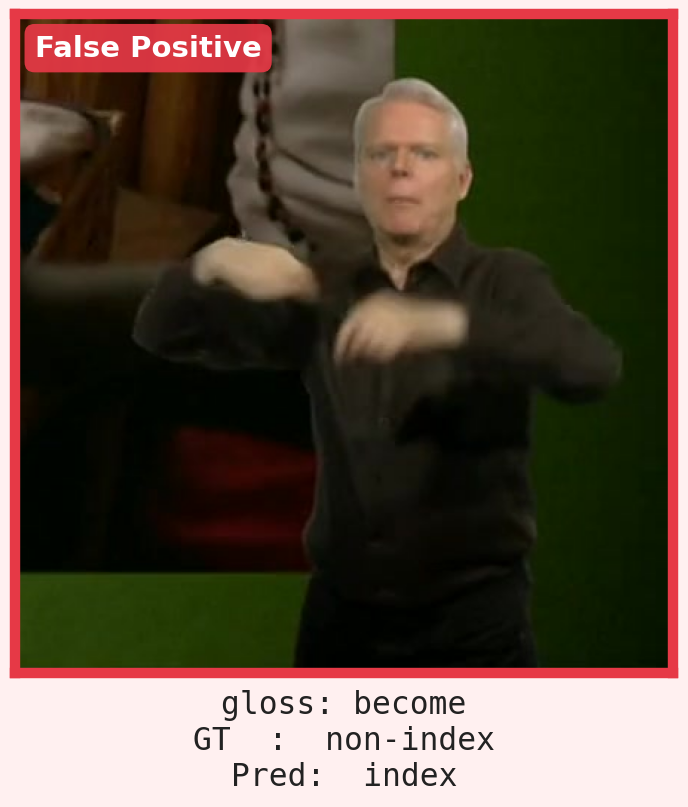} &
\includegraphics[width=0.24\textwidth]{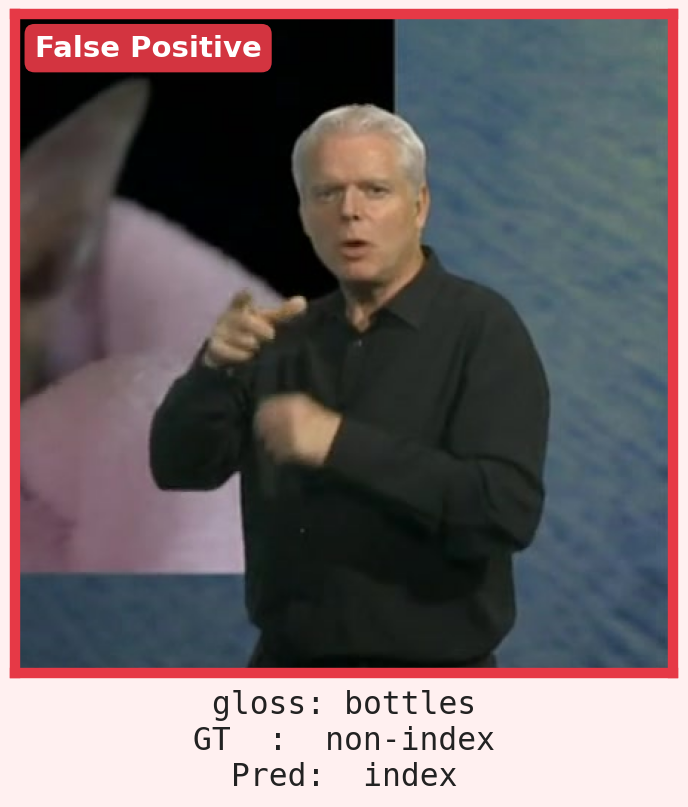} &
\includegraphics[width=0.24\textwidth]{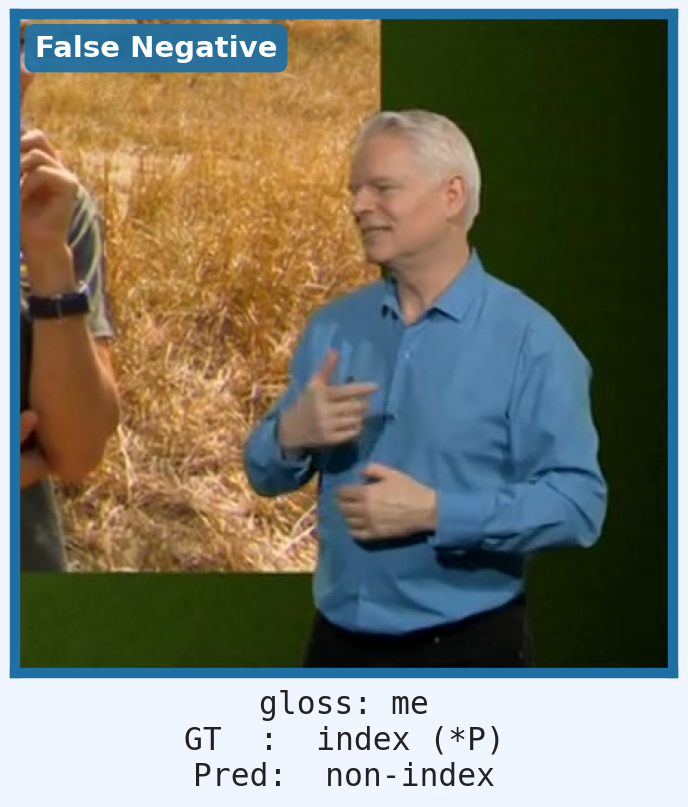} &
\includegraphics[width=0.24\textwidth]{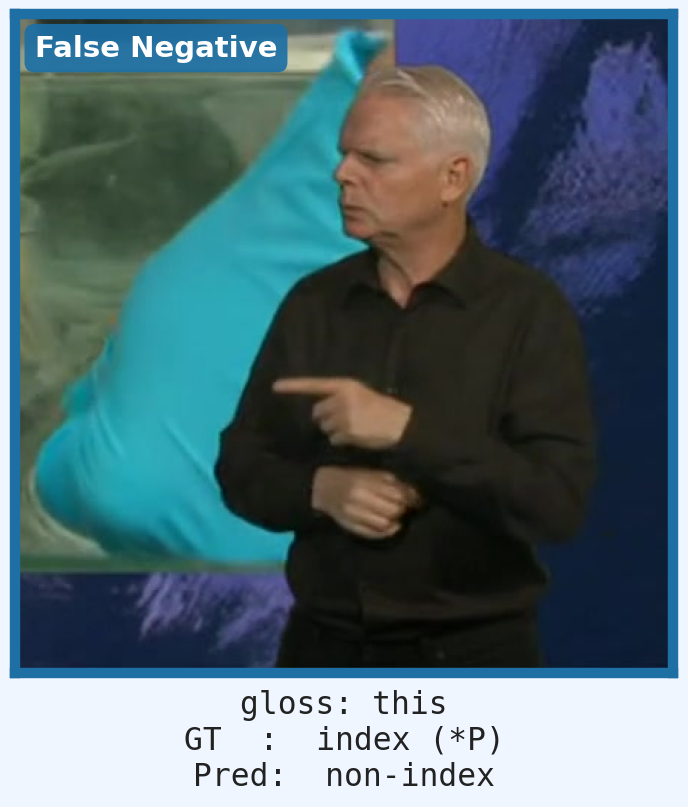} \\

{\tiny I} & {\tiny J} & {\tiny K} & {\tiny L} \\
\includegraphics[width=0.24\textwidth]{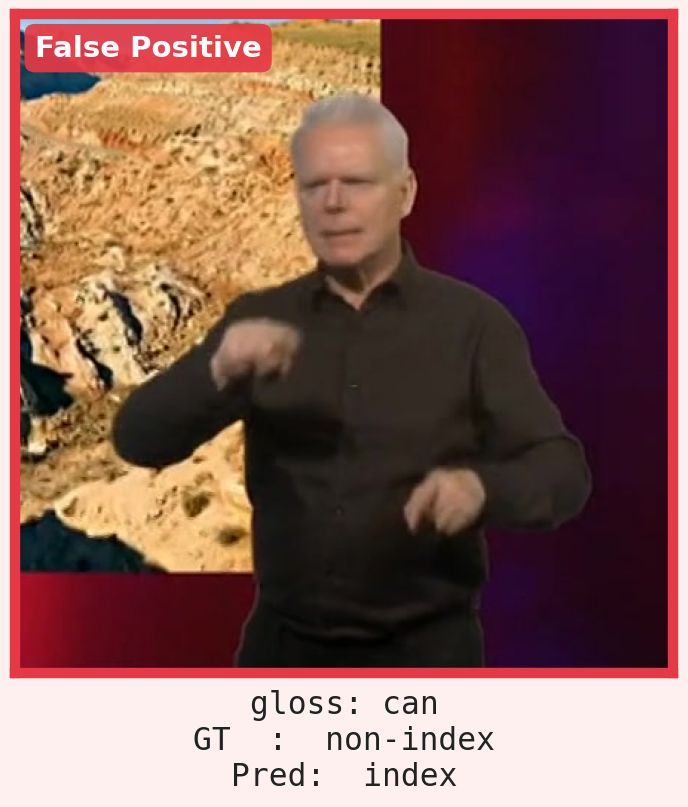} &
\includegraphics[width=0.24\textwidth]{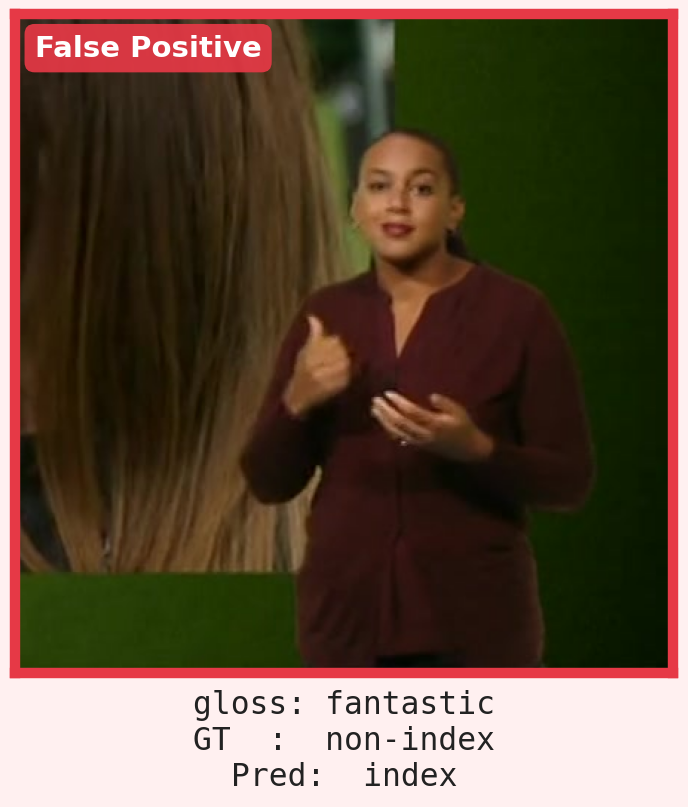} &
\includegraphics[width=0.24\textwidth]{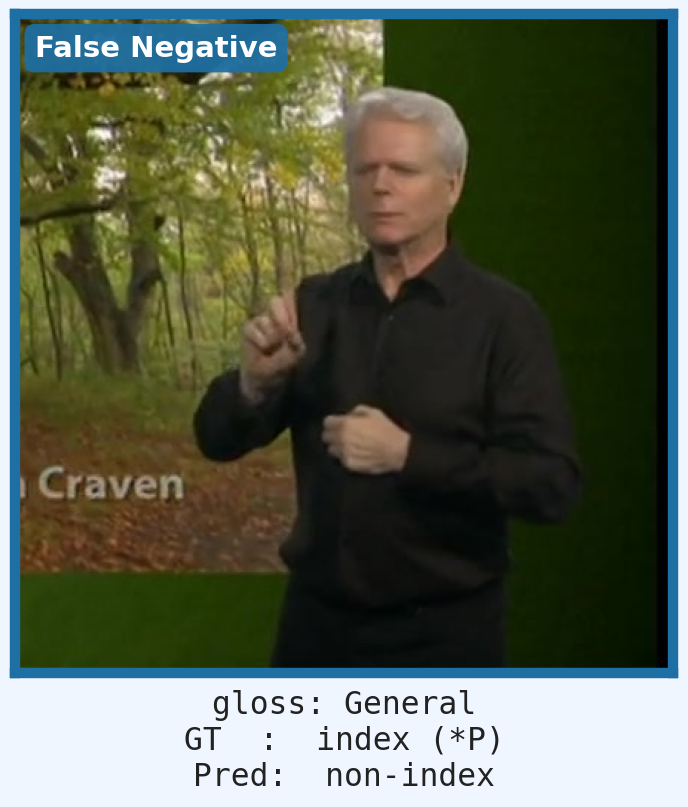} &
\includegraphics[width=0.24\textwidth]{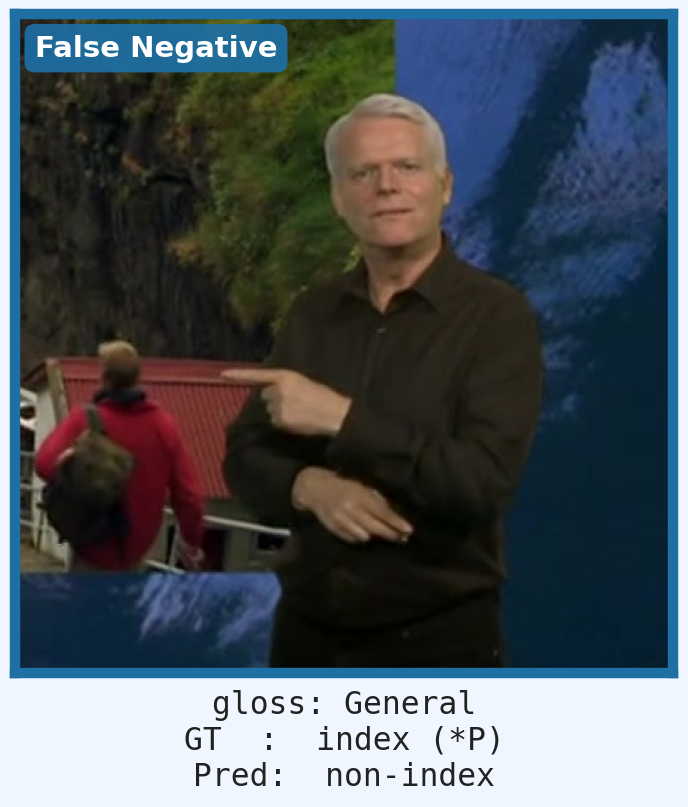} \\

\end{tabular}

\end{figure*}

We analyse the failure cases of the IPN
(trained on BSLCP and MDGS, evaluated on BOBSL with strict timestamp
matching) to assess whether misclassifications are systematic or
incidental. Figure \ref{fig:phase1_fp_fnb} shows representative failure
cases; each panel displays the centre frame of the segment, with border
colour indicating error type: \textcolor{red}{red} = FP (non-pointing
sign predicted as index), \textcolor{blue}{blue} = FN (pointing sign
missed).

The false positives are largely systematic: the mispredicted signs
frequently exhibit index-like properties, such as an extended or
body-directed handshape oriented toward a location where referential
pointing typically occurs. This is unsurprising, since the IPN operates
on skeletal motion alone and has no access to the discourse context that
would disambiguate a pointing gesture from a visually similar lexical
sign. 
The false negatives reveal a complementary failure mode: many missed
pointing signs are co-articulated with a co-occurring lexical sign,
producing a composite articulation whose skeletal trajectory is pulled
away from a canonical pointing gesture. Without access to wider
contextual features or an explicit model of co-articulation, the
network fails to isolate the indexical component within these
compound productions.

\section{Qualitative Results of ELM on BOBSL}
\label{sec:qual_phase2}
\begin{figure*}[!h]
    \centering
    \includegraphics[width=\linewidth]{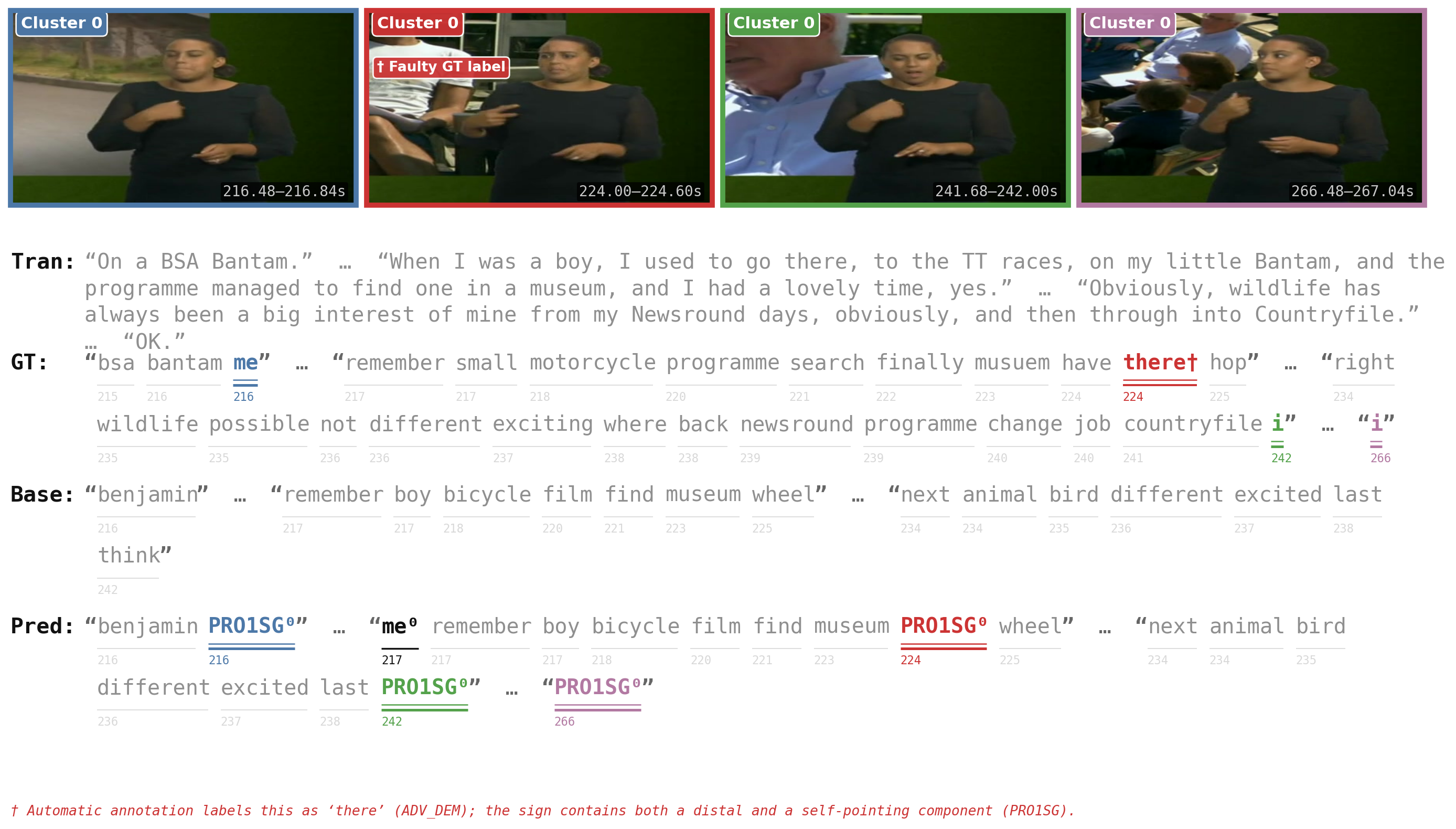}
    \caption{Instances of PRO1SG (\textit{me}/\textit{I}) across four examples in episode 6040895553921856506. Three instances are correctly detected and grouped. The second panel (red border) highlights an apparent ground-truth annotation or segmentation error: the sign contains both a locative (\textit{there}) and a self-pointing (\textit{me}) component, but the assigned label is only reflecting the former. The system nevertheless predicts PRO1SG consistently, suggesting that the pose features capture the self-referential component despite annotation noise.}
    \label{fig:cluster_vis_bobsl_a}
\end{figure*}

\begin{figure*}[!h]
    \centering
    \includegraphics[width=\linewidth]{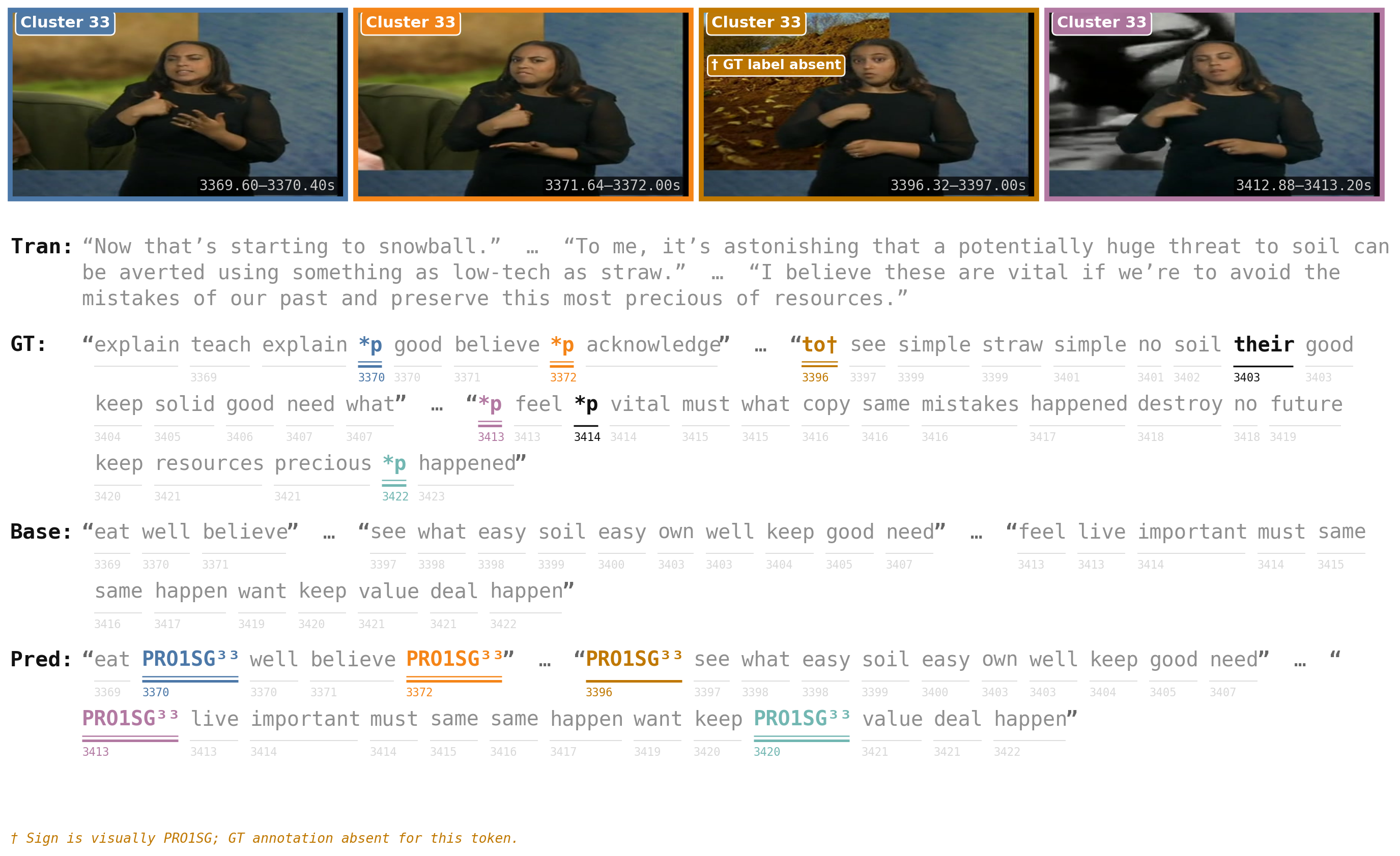}
    \caption{Instances of PRO1SG (\textit{me}) across four examples in episode 6003446875091426252. The third panel highlights a missing ground-truth annotation: the video contains a self-pointing sign but is glossed as the preposition \textit{to}, reflecting incomplete coverage of indexing signs. The model groups all four instances consistently and predicts PRO1SG throughout. The ground truth also reflects that two signs are not detected by either the baseline or the index expert.}
    \label{fig:cluster_vis_bobsl_b}
\end{figure*}

\begin{figure*}[!h]
    \centering
    \includegraphics[width=\linewidth]{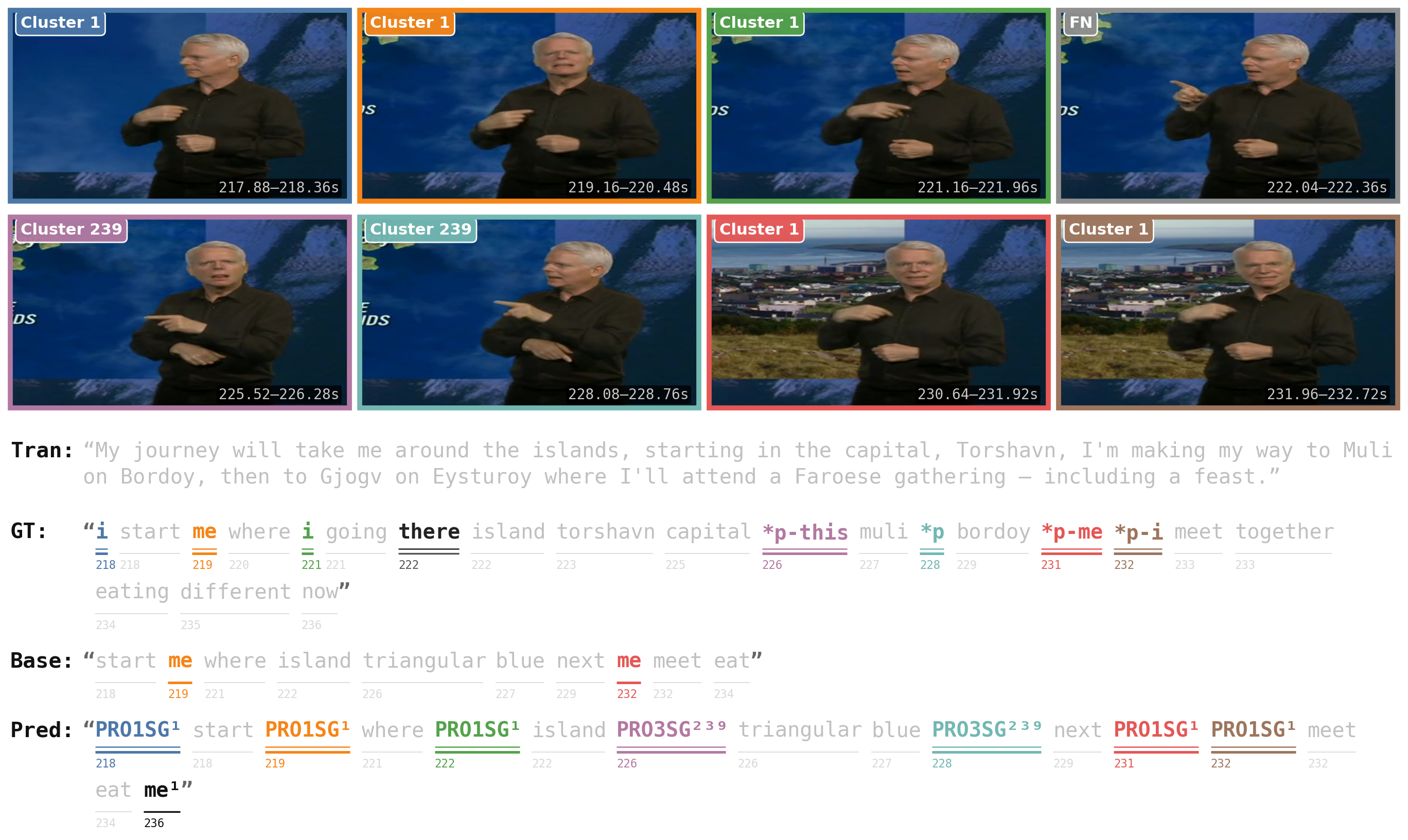}
    \caption{Two co-occurring entity clusters in episode 6177195911563690266. There are 5 instances of Cluster~1 (PRO1SG, \textit{me}/\textit{I}): all correctly detected and grouped. The fourth panel (grey border) is a false negative where a deictic \textit{there} token is missed. There are two instances of a PRO3SG cluster, a distinct third-person referent introduced in the same sentence window. This demonstrates that the system maintains both entities simultaneously and continues to recover the self-reference across a substantial temporal gap. The baseline predicts only two of the eight pointing tokens.}
    
    \label{fig:cluster_vis_bobsl_c}
\end{figure*}

Figures~\ref{fig:cluster_vis_bobsl_a}--\ref{fig:cluster_vis_bobsl_c} show predicted entity clusters on BOBSL, where only sparse referential labels are available. We show the predicted cluster membership alongside the subtitles for each segment, demonstrating that the cluster structure is reflected in coherent lexical output and visual similarity across instances assigned to the same entity.

\begin{figure*}[!t]
  \centering
  \includegraphics[width=\linewidth]{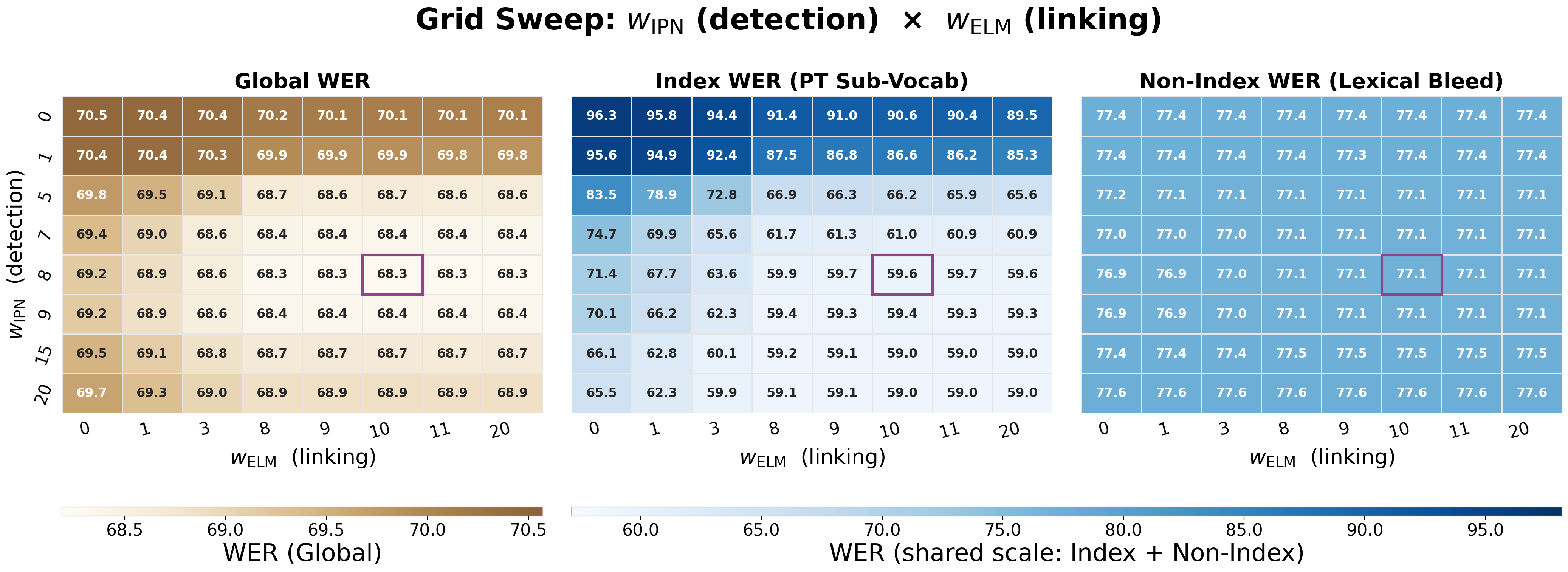}
  \caption{2D projection of the 3D heatmap showing numerical WER values per grid cell.}
  \label{fig:grid2d}
\end{figure*}

\newpage



\section{Numerical Values of SLR Grid Sweeps}
\label{app:sweeps}

Figure~\ref{fig:grid2d} displays the numerical WER values of each cell in the 3D heatmap. The cell chosen for optimal performance is highlighted with a purple border, corresponding to $w_{\text{IPN}} = 8$ and $w_{\text{ELM}} = 10$. The distribution reflects the underlying temperature parameter of CSLR2, which is set at $t = 20$.

\section{Automatic Entity Label Generation}
\label{sec:label_gen}

This section describes the two-stage pipeline used to generate entity labels
for the coreference training data (Sec.~5.2 of the main paper): mention
extraction, LLM cluster assignment, validation and retry, and the
post-processing step that resolves pointing-gesture glosses.

\subsubsection{Stage 1: Rule-Based Mention Candidate Extraction}
\label{app:mentions}

For each document, mention candidates are extracted from English free translations of MDGS and BSLCP using a spaCy POS-based span selector with \texttt{en\_core\_web\_lg}.  Four span patterns
are applied and merged:

\begin{enumerate}[noitemsep]
  \item \textbf{Proper-noun sequences:} one or more consecutive \texttt{PROPN}
        tokens (e.g.\ ``London'', ``British Sign Language'').
  \item \textbf{Nominal phrases:} optional \texttt{DET}/\texttt{PRON}/\texttt{NUM},
        followed by zero or more \texttt{ADJ} and one or more
        \texttt{NOUN}/\texttt{PROPN} tokens (e.g.\ ``the deaf community'').
  \item \textbf{Pronominal mentions:} single-token \texttt{PRON}
        (e.g.\ I, you, she, it).
  \item \textbf{Standalone demonstratives:} \texttt{DET} tokens from
        \{this, that, these, those\} not followed by a noun.
\end{enumerate}

Nested spans are pruned to maximal spans.  The output is one line per
dialogue turn with its extracted mentions:

\begin{tcolorbox}[colback=gray!5, fontupper=\scriptsize\ttfamily, boxrule=0.4pt]
\begin{verbatim}
[turn=3 speaker=BF25F29WHN] "So I started school in London",
    ["I", "school", "London"]
\end{verbatim}
\end{tcolorbox}

\subsubsection{Stage 2: LLM Coreference Cluster Assignment}
\label{app:llm}

\paragraph{Model setup.}
We use a 120B-parameter open-weights model (\texttt{gpt-oss:120b}) via Ollama
(\texttt{/api/chat}, \texttt{temperature=0}, 8K-token context, max 2K
generation tokens; extended to 16K on \texttt{done\_reason=length}).

\paragraph{Prompt design.}
The system prompt specifies task constraints and output format; the user
message provides the document identifier and formatted dialogue.
\begin{tcolorbox}[title=System Prompt,
                  fontupper=\scriptsize,
                  breakable,
                  listing only,
                  listing options={language={}, basicstyle=\ttfamily\scriptsize, breaklines=true}]
You are a coreference clustering engine.

INPUT
Each line is:
[turn=T speaker=S] "TURN TEXT", ["mention1", "mention2", ...]

TASK
For each mention in the list, assign a coreference cluster ID (an integer).

CORE RULES (PROCESSING ORDER)

1) Process the dialogue strictly in order, from the first turn to the last.

2) When a mention refers to an entity already seen, reuse the SAME cluster ID.

3) When a mention is new, assign the NEXT unused integer (starting at 0).

4) Never renumber, reuse, or reshuffle existing cluster IDs.

5) If a mention is not an entity mention or you are unsure, output "-".

REFERENCE RULES
- Same entity => same integer.
- I / me / my => current speaker.
- you / your => most likely addressee.

OUTPUT FORMAT
- Return ONLY one bracketed list per input line.
- Item count MUST equal mention count.
Examples: [0]   [1, 2, 3]   [4, 3, -]
\end{tcolorbox}

\begin{tcolorbox}[title=User Message Format,
                  colback=gray!5,
                  fontupper=\scriptsize,
                  breakable,
                  listing only,
                  listing options={basicstyle=\ttfamily\scriptsize, breaklines=true}]
doc\_id: BF25F29WHN

Dialogue:

[turn=1 speaker=BF25F29WHN] "Well I was born deaf", ["I"]

[turn=2 speaker=BF25F29WHN] "Right and where did you grow up?", ["you"]

...
\end{tcolorbox}

\paragraph{Document chunking.}
Documents are split into non-overlapping \textbf{100-turn chunks} processed
sequentially.  IDs from chunk $k$ are offset by $\max(\text{IDs in
chunks }1\ldots k{-}1)+1$ before concatenation to ensure globally unique
cluster IDs.

\paragraph{Validation and retry.}
Each LLM response is checked for (i) correct line count and (ii) correct
per-line item count.  Failures trigger up to \textbf{three retries}; if all
fail and the chunk has $\geq$20 turns, it is recursively \textbf{halved}
(minimum 10 turns) and each half retried.  Successful chunks are cached for
resume. 

\begin{table*}[!b]
\caption{Mapping between graph nodes and SMPL-X body joints.}
\label{tab:smplx_mapping}
\centering
\small
\setlength{\tabcolsep}{4pt}
\begin{tabular}{@{}lcccccccc@{}}
\toprule
\textbf{Node} & 0 & 1 & 2 & 3 & 4 & 5 & 6 & 7 \\
\midrule
\textbf{SMPL-X idx} & 21 & 19 & 17 & 16 & 18 & 20 & 0 & 22 \\
\textbf{Joint (side)} & Right wrist & Right elbow & Right shoulder & Left shoulder & Left elbow & Left wrist & Pelvis & Upper spine \\
\bottomrule
\end{tabular}
\end{table*}

\subsubsection{Stage 3: Resolving Pointing-Gesture Glosses}
\label{app:ptresolver}

The LLM assigns clusters to lexical mentions in the translation.  Sign-language
corpora additionally contain pointing-gesture glosses absent from the
translation; these are resolved in post-processing:

\begin{enumerate}[nosep]
  \item \textbf{Explicit mapping.}  Fixed-referent glosses are resolved
        deterministically (\texttt{PT:PRO1SG}$\to$first-person cluster;
        \texttt{PT:PRO2SG}$\to$second-person cluster; etc.).
  \item \textbf{Generic pointers.}  All bare \texttt{PT:} tokens are first
        collapsed into a single shared cluster.  \texttt{PT:} and \texttt{PT:DET}
        tokens are then grouped into runs; temporally proximate occurrences
        (within 25 entries) are merged into the same cluster.
  \item \textbf{Soft third-person lookup.}  Before co-occurrence scoring,
        \texttt{PT:PRO3SG}, \texttt{PT:POSS3SG}, \texttt{PT:PRO3PL}, and
        \texttt{PT:POSS3PL} labels search \texttt{entities\_raw} for clusters
        containing a third-person pronoun mention
        (\textit{he, she, her, his, it, they, them, their}),
        providing a lightweight resolution path that avoids spurious co-occurrence matches.
  \item \textbf{Cross-sentence co-occurrence.}  Remaining PT labels are linked
        to the LLM cluster with which they most frequently co-occur (processed
        by decreasing cluster size).
  \item \textbf{BUOY grouping.}  Nearby buoy tokens
        (\texttt{PT:BUOY}, \texttt{PT:LBUOY}, \texttt{PT:FBUOY}) within a
        15-entry window share a cluster.
  \item \textbf{Fallback.}  Remaining instances attach to the nearest
        assigned cluster of the same label type, or open a new cluster.
\end{enumerate}

\subsubsection{Stage 4: Rule-Based Post-Processing Refinements}
\label{app:postproc}

Raw LLM clusters often contain systematic errors that degrade training label quality. Three targeted refinements are applied after Stage~3:

\paragraph{Inanimate co-referent suppression.}
A regex pattern blocks inanimate temporal, weather, and event nouns
(e.g.\ \textit{a day}, \textit{the snow}, \textit{the wedding}) from being
assigned to person-pointing clusters. The pattern is applied at two levels:
(i) cluster-level, where any cluster whose representative mention matches the
pattern is excluded from matching against person-pointing PT labels; and
(ii) mention-level, where per-sentence inanimate mentions are skipped when
resolving person-pointing labels. This corrects a systematic LLM error where
temporal or locative noun phrases are grouped into the same cluster as a
person due to surface proximity in the text.

\paragraph{Person-label cluster unification.}
After co-occurrence-based assignment, the same person-pointing label (e.g.\
\texttt{PT:PRO3SG}) may be fragmented across several small clusters within a
document. A second pass identifies all clusters assigned the same
person-pointing label and merges them toward the cluster with the strongest
evidence, updating all mention-level assignments accordingly.

\paragraph{Possessive cluster merging.}
Clusters where all mentions match a possessive relation pattern
(e.g.\ \textit{my wife}, \textit{my girlfriend}) and all are assigned the
same person-pointing label are merged under the dominant entity. This
addresses cases where possessively-introduced referents are fragmented
across multiple clusters due to variation in surface form.

\subsubsection{Constructing a Small Evaluation Suite}
\label{app:evalsuite}

To obtain a coarse estimate of the performance of the LLM-based
coreference pipeline, we constructed a small manually annotated
evaluation set. For three documents from BSLCP, we created
ground-truth entity-link annotations by hand.

While these annotations were produced by the authors themselves and
therefore cannot be considered perfect, they provide a reasonable
approximation for comparing different model configurations.
In practice, the annotation process scales reasonably well,
requiring approximately five minutes per document.

Using this setup, we evaluated several LLM-based coreference
pipelines running on the DGS Spark cluster via Ollama.
Preliminary experiments showed that most models failed to produce
stable or well-structured cluster outputs. In practice,
\texttt{gpt-oss:120b} was the only free-to-use model that consistently produced
usable results, and it was therefore selected for the large-scale
annotation pipeline. Table~\ref{tab:coref_eval} reports the accuracy of several tested
models on the three annotated documents.
\begin{table}[tb]
\caption{Accuracy of LLM-based coreference annotations on three manually annotated documents.}
\label{tab:coref_eval}
\centering
\small
\begin{tabular}{@{}lccc@{}}
\toprule
\textbf{Model} & \textbf{Acc.\ Doc 1} & \textbf{Acc.\ Doc 2} & \textbf{Acc.\ Doc 3} \\
\midrule
ChatGPT 5.2  & 0.82 & 0.88 & 0.85 \\
gpt-oss:120b & 0.76 & 0.93 & 0.85 \\
gpt-oss:20b  & 0.63 & 0.83 & 0.78 \\
\bottomrule
\end{tabular}
\end{table}

\section{Input Feature Layout and ELM Auxiliary Features}
\label{sec:feature_layout}

Our model operates on pose representation from two 3D keypoint sources, combined into a single skeletal graph processed by the SL-GCN backbone. 
\paragraph{WiLoR Hand Keypoints.}
WiLoR produces 21 MANO-compatible 3D keypoints per hand (wrist +
$5\!\times\!4$ finger joints).  Both hands are included, giving
$2\!\times\!21 = 42$ additional nodes (right hand: nodes 8–28; left hand:
nodes 29–49).

\begin{figure*}[!b]
  \centering
  \includegraphics[width=\linewidth]{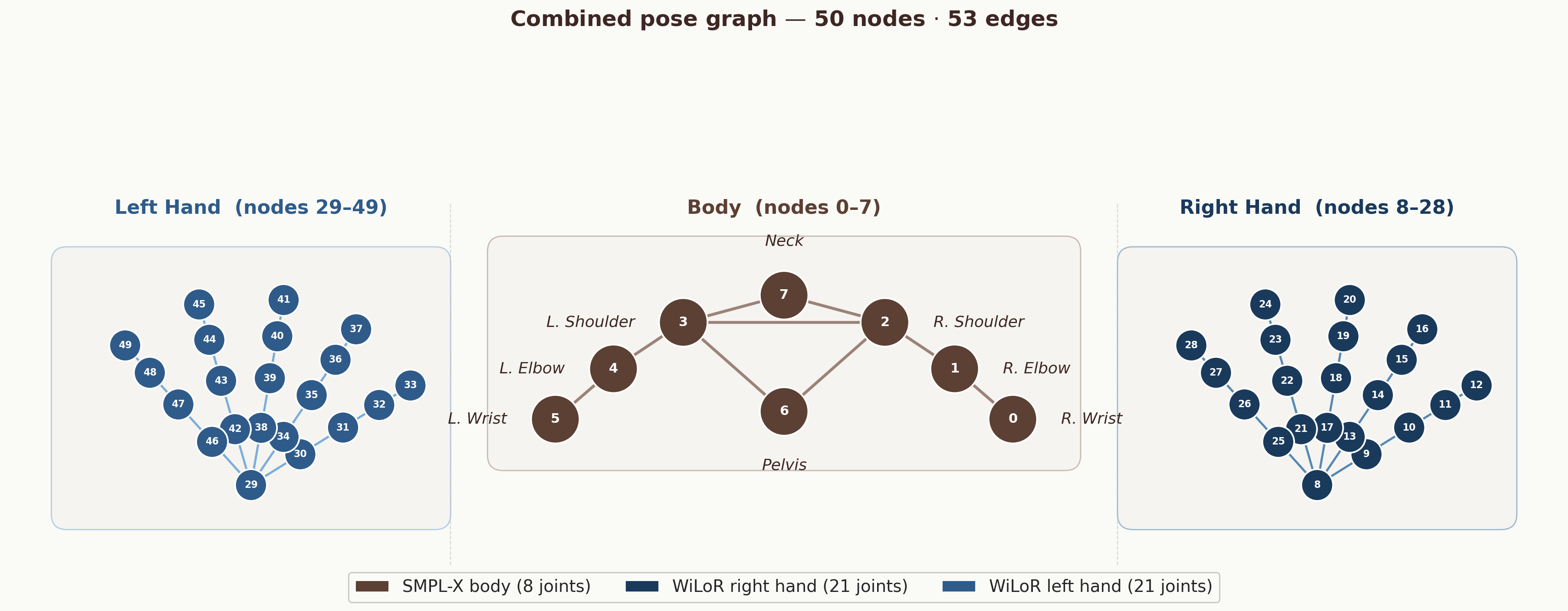}
  \caption{Combined pose graph.  Brown: 8 SMPL-X body joints (nodes 0–7).
           Dark blue / steel blue: WiLoR right (8–28) and left (29–49) hands,
           21 joints each: $
  \mathcal{V} = \bigl\{V_{\mathrm{body}}^{0\ldots7},\;
                       V_{\mathrm{R}}^{8\ldots28},\;
                       V_{\mathrm{L}}^{29\ldots49}\bigr\}.
$}
  \label{fig:pose_graph}
\end{figure*}

\paragraph{SMPL-X Body Keypoints.}
Whole-body poses are estimated from RGB video using SMPL-X, which
outputs 55 joints per detected person.  We retain the 8 joints that are informative towards the relative wrist position to body and motion trajectories. The map between the graph nodes and the SMPL-X joints are given in Table \ref{tab:smplx_mapping}.

\paragraph{Combined Graph.}
The full graph $\mathcal{G}=(\mathcal{V},\mathcal{E})$ has $|\mathcal{V}|=50$
nodes and $|\mathcal{E}|=53$ undirected edges as depicted in figure \ref{fig:pose_graph}.

\paragraph{Normalisation Pipeline.}
Each node carries raw 3D coordinates $(x,y,z)$; the input tensor per temporal
window is $(T\!\times\!50\!\times\!3)$ with $T{=}12$ frames.  Body-joint coordinates lie approximately in $[-1,1]$; hand joints are scaled
proportionally. Before use,
coordinates undergo:
\begin{enumerate}[nosep]
  \item \textbf{Scale alignment.}  Body skeleton auto-scaled so that the
        mean body–hand limb length matches a fixed target of $0.075$ in WiLoR
        metric units, preventing the larger SMPL-X range from dominating hand
        features.
  \item \textbf{Wrist stitching.}  Body wrist nodes (0, 5) are translated to
        coincide with their WiLoR counterparts, enforcing a consistent hand
        origin.
  \item \textbf{Root centring.}  Skeleton translated so that the mean wrist
        position over the window lies at the origin.
  \item \textbf{Scale normalisation.}  All coordinates divided by the
        inter-shoulder distance (nodes 2–3), giving viewpoint invariance.
  \item \textbf{Canonical orientation.}  Rotation aligns the
        shoulder–shoulder axis with the $y$-axis, removing in-plane torso
        rotation while preserving relative hand pose.
\end{enumerate}

\subsection{ELM Auxiliary Geometric Features}
\label{app:elm_features}

As described in Sec.~4.3 of the main paper, the ELM scoring incorporates auxiliary feature streams that recover spatial and kinematic information compressed away by the IPN binary classification.

\paragraph{Engineered pointing scalar features.}
Six scalar features are computed from the dominant pointing hand (selected as
the hand with greater mean wrist--elbow displacement over the clip):

\begin{enumerate}[nosep]
  \item \textbf{Elevation.} Vertical angle of the mean pointing direction
        $\arcsin(d_y)$, where $\mathbf{d}$ is the normalised sum of the finger
        vector (tip$-$wrist) and arm vector (wrist$-$elbow).
  \item \textbf{Target-$y$.} Mean fingertip height relative to the shoulder
        midpoint, normalised by shoulder width.  Distinguishes upward (addressee,
        abstract loci) from body-level (self, nearby referents) pointing.
  \item \textbf{Target-$z$.} Mean fingertip depth relative to the shoulder
        midpoint, normalised by shoulder width.  Distinguishes forward-directed
        (third-person, distal) from body-proximal pointing.
  \item \textbf{Arm reach.} Normalised elbow-to-fingertip distance, capturing
        how fully extended the arm is.
  \item \textbf{Index selectivity.} Difference between the normalised index
        fingertip extension and the mean extension of the middle, ring, and
        pinky fingertips, measured relative to palm scale.  High values
        indicate a prototypical pointing handshape.
  \item \textbf{Trajectory length.} Cumulative fingertip displacement over the
        clip, normalised by shoulder width.  Distinguishes static holds from
        sweeping or arc-shaped pointing gestures.
\end{enumerate}

For scoring compatibility with entity $e_k$, the six values for the current
mention are concatenated with the stored per-cluster running mean and their
element-wise difference (18 values total), then projected to a fixed-size
embedding via a two-layer MLP with LayerNorm.

\section{IPN Threshold Ablation}
\label{sec:threshold}

Table~\ref{tab:threshold_sweep} reports IPN performance as a function
of the decision threshold $\tau$ (B+M joint model, 3 seeds).  The operating
point $\tau{=}0.70$ maximises F1 on the index class.  For SLR integration we
use $\tau{=}0.90$ to further suppress false positives at the cost of recall,
trading lexical bleed for precision in the downstream logit boost.

\begin{table}[!h]
  \small
  \centering
\setlength{\tabcolsep}{4pt}
  \begin{tabular}{@{}ccccccc@{}}
    \toprule
    $\tau$ & F\% & Prec-I & Rec-I & \textbf{F1-I} & BalAcc \\
    \midrule
    0.30 & $28$ & $0.50_{\pm0.04}$ & $0.89_{\pm0.02}$ & $0.64_{\pm0.03}$ & $0.86_{\pm0.01}$ \\
    0.35 & $26$ & $0.53_{\pm0.04}$ & $0.88_{\pm0.03}$ & $0.66_{\pm0.03}$ & $0.87_{\pm0.01}$ \\
    0.40 & $25$ & $0.56_{\pm0.04}$ & $0.86_{\pm0.03}$ & $0.68_{\pm0.03}$ & $0.87_{\pm0.01}$ \\
    0.45 & $23$ & $0.58_{\pm0.04}$ & $0.84_{\pm0.03}$ & $0.68_{\pm0.02}$ & $0.86_{\pm0.01}$ \\
    0.50 & $22$ & $0.60_{\pm0.03}$ & $0.82_{\pm0.03}$ & $0.69_{\pm0.02}$ & $0.86_{\pm0.01}$ \\
    0.55 & $21$ & $0.62_{\pm0.03}$ & $0.80_{\pm0.02}$ & $0.70_{\pm0.02}$ & $0.85_{\pm0.01}$ \\
    0.60 & $19$ & $0.64_{\pm0.03}$ & $0.78_{\pm0.02}$ & $0.70_{\pm0.01}$ & $0.85_{\pm0.01}$ \\
    0.65 & $18$ & $0.67_{\pm0.03}$ & $0.75_{\pm0.02}$ & $0.71_{\pm0.01}$ & $0.84_{\pm0.01}$ \\
    0.70 & $16$ & $0.69_{\pm0.02}$ & $0.72_{\pm0.03}$ & $\mathbf{0.71}_{\pm0.01}$ & $0.83_{\pm0.01}$ \\
    0.75 & $15$ & $0.73_{\pm0.02}$ & $0.68_{\pm0.02}$ & $0.70_{\pm0.01}$ & $0.82_{\pm0.01}$ \\
    0.80 & $13$ & $0.76_{\pm0.02}$ & $0.63_{\pm0.04}$ & $0.69_{\pm0.02}$ & $0.80_{\pm0.02}$ \\
    0.85 & $11$ & $0.80_{\pm0.02}$ & $0.56_{\pm0.04}$ & $0.66_{\pm0.02}$ & $0.77_{\pm0.02}$ \\
    0.90 & $9$  & $0.84_{\pm0.02}$ & $0.45_{\pm0.07}$ & $0.58_{\pm0.05}$ & $0.72_{\pm0.03}$ \\
    0.95 & $5$  & $0.89_{\pm0.03}$ & $0.28_{\pm0.08}$ & $0.42_{\pm0.10}$ & $0.64_{\pm0.04}$ \\
    \bottomrule
  \end{tabular}
  \caption{Threshold sweep (B+M joint model, mean $\pm$ std, 3 seeds).
           Best F1-I bolded. F \% Denotes the percent of indexing clips fiered. }
  \label{tab:threshold_sweep}
\end{table}

Table~\ref{tab:threshold_sweep_detail} further reports the cw4-r1 model
across individual evaluation datasets at representative thresholds.

\begin{table*}[!t]
\caption{Threshold sweep for cw4-r1 across evaluation datasets
         (B = BSLCP, M = MDGS, Bob$_{fs}$ = BOBSL full-sign).}
\label{tab:threshold_sweep_detail}
\centering
\setlength{\tabcolsep}{4pt}
\small
\begin{tabular}{@{}llccccccc@{}}
\toprule
Train & Eval & $\tau$ & BalAcc & Macro F1 & Prec-L & Rec-L & Prec-I & Rec-I \\
\midrule
cw4-r1 & B
  & 0.60 & $0.80_{\pm0.00}$ & $0.80_{\pm0.00}$ & $0.74_{\pm0.00}$ & $0.94_{\pm0.00}$ & $0.92_{\pm0.00}$ & $0.66_{\pm0.00}$ \\
  & & 0.70 & $0.79_{\pm0.00}$ & $0.79_{\pm0.00}$ & $0.73_{\pm0.00}$ & $0.95_{\pm0.00}$ & $0.92_{\pm0.00}$ & $0.64_{\pm0.00}$ \\
  & & 0.80 & $0.78_{\pm0.00}$ & $0.77_{\pm0.00}$ & $0.71_{\pm0.00}$ & $0.96_{\pm0.00}$ & $0.94_{\pm0.00}$ & $0.60_{\pm0.00}$ \\
  & & 0.90 & $0.75_{\pm0.00}$ & $0.74_{\pm0.00}$ & $0.68_{\pm0.00}$ & $0.97_{\pm0.00}$ & $0.95_{\pm0.00}$ & $0.54_{\pm0.00}$ \\
\midrule
cw4-r1 & M
  & 0.60 & $0.85_{\pm0.00}$ & $0.84_{\pm0.00}$ & $0.80_{\pm0.00}$ & $0.92_{\pm0.00}$ & $0.91_{\pm0.00}$ & $0.77_{\pm0.00}$ \\
  & & 0.70 & $0.84_{\pm0.00}$ & $0.83_{\pm0.00}$ & $0.78_{\pm0.00}$ & $0.93_{\pm0.00}$ & $0.92_{\pm0.00}$ & $0.74_{\pm0.00}$ \\
  & & 0.80 & $0.82_{\pm0.00}$ & $0.81_{\pm0.00}$ & $0.75_{\pm0.00}$ & $0.94_{\pm0.00}$ & $0.93_{\pm0.00}$ & $0.69_{\pm0.00}$ \\
  & & 0.90 & $0.78_{\pm0.00}$ & $0.78_{\pm0.00}$ & $0.71_{\pm0.00}$ & $0.96_{\pm0.00}$ & $0.94_{\pm0.00}$ & $0.61_{\pm0.00}$ \\
\midrule
cw4-r1 & Bob$_{fs}$
  & 0.60 & $0.68_{\pm0.00}$ & $0.71_{\pm0.00}$ & $0.90_{\pm0.00}$ & $0.96_{\pm0.00}$ & $0.63_{\pm0.00}$ & $0.41_{\pm0.00}$ \\
  & & 0.70 & $0.68_{\pm0.00}$ & $0.71_{\pm0.00}$ & $0.90_{\pm0.00}$ & $0.96_{\pm0.00}$ & $0.65_{\pm0.00}$ & $0.39_{\pm0.00}$ \\
  & & 0.80 & $0.67_{\pm0.00}$ & $0.70_{\pm0.00}$ & $0.89_{\pm0.00}$ & $0.97_{\pm0.00}$ & $0.67_{\pm0.00}$ & $0.37_{\pm0.00}$ \\
  & & 0.90 & $0.66_{\pm0.00}$ & $0.69_{\pm0.00}$ & $0.89_{\pm0.00}$ & $0.97_{\pm0.00}$ & $0.69_{\pm0.00}$ & $0.34_{\pm0.00}$ \\
\bottomrule
\end{tabular}
\end{table*}


\section{WER$_{all}$, WER$_{Index}$ and WER$_{Lex}$}
\label{sec:wer_metrics}

\paragraph{Word Error Rate (WER).}
WER is the Levenshtein edit distance between predicted sequence $\hat{Y}$ and
reference $Y$, normalised by reference length:
\begin{equation}
  \mathrm{WER} = \frac{S + D + I}{N} \times 100\%,
  \label{eq:wer}
\end{equation}
where $S$, $D$, $I$ are substitutions, deletions, and insertions, and
$N=|Y|$.  Evaluation is case-insensitive and synonym-aware
(Sec.~\ref{app:synonyms}).

\paragraph{WER$_{Index}$.}
Both reference and hypothesis are filtered to the PT sub-vocabulary
$\mathcal{V}_{\mathrm{PT}}$ (pronouns, demonstratives, reflexives) before
applying Eq.~\eqref{eq:wer}:
\begin{equation}
  \mathrm{WER_{Index}} =
    \frac{\mathrm{EditDist}\!\left(
      \hat{Y}|_{\mathcal{V}_{\mathrm{PT}}},\;
      Y|_{\mathcal{V}_{\mathrm{PT}}}
    \right)}
    {\bigl|Y|_{\mathcal{V}_{\mathrm{PT}}}\bigr|}
    \times 100\%.
\end{equation}

\paragraph{WER$_{Lex}$.}
All $\mathcal{V}_{\mathrm{PT}}$ tokens are \emph{excluded} prior to alignment:
\begin{equation}
  \mathrm{WER_{Lex}} =
    \frac{\mathrm{EditDist}\!\left(
      \hat{Y}|_{\overline{\mathcal{V}_{\mathrm{PT}}}},\;
      Y|_{\overline{\mathcal{V}_{\mathrm{PT}}}}
    \right)}
    {\bigl|Y|_{\overline{\mathcal{V}_{\mathrm{PT}}}}\bigr|}
    \times 100\%.
\end{equation}
This metric is unaffected by pointing-sign improvements and serves as a
sanity check that the base model is not degraded due to lexical bleed.

\section{Synonym-Based Evaluation Mapping}
\label{app:synonyms}
Several pronoun and demonstrative forms share the same BSL sign (eg.
\textit{I} and \textit{me} are both realised as a chest-point).  To avoid
penalising semantically correct but lexically variant predictions, synonymous
forms within each group are treated as equivalent during alignment. A different challenge is that several out-of-vocabulary (OOV) forms appear in the test set but not in the pre-trained model vocabulary. To include these in the evaluation, we map these forms to their in-vocabulary closest visual and semantic neighbor before scoring (Table~\ref{tab:synonym_mapping}).

\begin{table}[!h]
  \centering
  \setlength{\tabcolsep}{6pt}
  \caption{BSL synonym groups.  OOV forms are mapped to the in-vocabulary
           synonym before WER alignment; ``---'' = no OOV forms.}
  \label{tab:synonym_mapping}
  \small
  \begin{tabular}{lll}
    \toprule
    \textbf{Group} & \textbf{In-vocabulary} & \textbf{OOV $\to$ mapped} \\
    \midrule
    PRO1SG  & me              & I                                   \\
    PRO2SG  & you             & ---                                 \\
    PRO3SG  & he, her         & him, she, it                        \\
    PRO1PL  & our           & we, us \\
    PRO3PL  & they            & them                                \\
    POS1SG  & my              & mine                                \\
    POS2SG  & your            & yours                               \\
    POS3SG  & her             & his, its                            \\
    POS1PL &  our             & ours           \\
    POS3PL  & their           & ---                                 \\
    DET\_SG   & this, that      & ---                                 \\
    ADV\_DEM  & here, there     & these, those                        \\
    REFL1SG & myself          & ---                                 \\
    REFL    & yourself        & himself, herself, \\ & &itself, themselves \\
    \bottomrule
  \end{tabular}
\end{table}

\section{SLR Integration: Inference-Time Guardrails}
\label{app:guardrails}

The IPN and ELM boosts are applied additively at inference time. Two
additional guardrails are employed to prevent false or inconsistent boosts from
degrading non-indexing predictions.

\paragraph{Directional consistency filter.}
At linking time, a candidate entity assignment is rejected if the cosine
similarity between the current mention's pointing direction and the stored mean
direction of the candidate cluster falls below a threshold
$\delta_{\text{dir}}$. In this case the mention is either assigned to a
different cluster or opens a new one. This prevents the cluster prior of a
left-side referent from being applied to a right-directed pointing sign, which
would add incorrect vocabulary bias. The threshold $\delta_{\text{dir}}=0.20$
is used in all reported experiments.

\paragraph{Cluster group partitioning.}
BSLCP PT gloss labels span multiple grammatical categories: person-pointing
pronouns (\texttt{PRO1SG}, \texttt{PRO2SG}, \texttt{PRO3SG}, etc.),
possessives (\texttt{POSS*}), locatives (\texttt{LOC}), determiners
(\texttt{DET}), and spatial buoys (\texttt{BUOY*}). A single entity cluster
accumulating evidence from both person-pointing and locative pointing would
produce a confused vocabulary prior. With cluster group partitioning enabled,
each cluster is restricted to one PT grammatical category: a mention of a
different category cannot be linked to an existing cluster of a different
group, forcing a new cluster to be opened instead. This ensures that
person-pointing and locative-pointing referents are tracked separately, and
that their respective vocabulary priors remain category-consistent.


\section{Hyperparameters}
\label{app:hyperparameters}
This section lists the hyperparameters used for training in IPN (Table~\ref{tab:hyp_phase1}), ELM (Table~\ref{tab:hyp_phase2}), and SLR integration (Table~\ref{tab:hyp_phase3}).

\begin{table}[!h]
\centering
\caption{IPN hyperparameters (Optuna best)}
\label{tab:hyp_phase1}
\scriptsize
\begin{tabular}{lll}
\toprule
\textbf{Hyperparameter} & \textbf{Value} & \textbf{Notes} \\
\midrule
Backbone            & SL-GCN         & Spatial-GCN on graph \\
Clip length         & 12 frames      & \\
Body keypoints      & 8 (SMPL-X)     & \\
Hand keypoints      & 42 (WiLoR, 21/hand) & \\
Batch size          & 96             & \\
Epochs              & 25             & Early stop patience: 7 \\
Learning rate       & $1.10\!\times\!10^{-3}$ & Adam \\
Weight decay        & $3.02\!\times\!10^{-4}$ & \\
Non-PT CW & 4.0            & CE loss upweight \\
Focal $\gamma$      & 0.0            & CE \\
Batch balance       & \texttt{WeightedRandomSampler} & 1:1 PT:non-PT \\
\bottomrule
\end{tabular}
\end{table}

\begin{table}[!h]
\centering
\caption{ELM hyperparameters.  Encoder and classification head
         are frozen.}
\label{tab:hyp_phase2}
\small
\begin{tabular}{ll}
\toprule
\textbf{Hyperparameter} & \textbf{Value} \\
\midrule
Encoder             & Frozen IPN SL-GCN  \\
Memory type         & \texttt{learned\_avg}, Hadamard similarity \\
Memory MLP size     & 300              \\
Memory MLP depth    & 2                \\
Entity slots        & 20               \\
Entity dim          & 10               \\
Linking proj.\ hidden & 256           , 2-layer MLP, GELU, LN \\
Dropout             & 0.3              \\
Batch size          & 32               \\
Epochs              & 10               \\
Learning rate       & $2.53\!\times\!10^{-4}$, Adam \\
Weight decay        & $8.07\!\times\!10^{-6}$ \\
$\lambda_\text{cls}$ & 0.1            , Aux. classification weight \\

\bottomrule
\end{tabular}
\end{table}

\begin{table}[!h]
\centering
\caption{CSLR2 inference parameters (no training; selected by sweep).}
\label{tab:hyp_phase3}
\small
\begin{tabular}{ll}
\toprule
\textbf{Hyperparameter} & \textbf{Value, note}  \\
\midrule
Expert threshold $\tau$       & 0.9   , IPN fire threshold \\
Scoring mode        & Window-level    ,  Single score per subtitle \\
Boost mode          & Soft (additive) , Non-PT logits unchanged \\
BOBSL frame rate    & 25\,fps         \\
Pointing token      & \texttt{"point"}, Vocabulary anchor \\
\bottomrule
\end{tabular}
\end{table}

\begin{table*}[!h]
\centering
\small
\setlength{\tabcolsep}{6pt}
\renewcommand{\arraystretch}{1.15}

\begin{tabular}{p{3.8cm} p{5.2cm} p{6.2cm}}
\hline
\textbf{Component} & \textbf{Architecture} & \textbf{Parameters} \\
\hline

IPN (Index Pointing Network, Phase 1) &
ST-GCN backbone + linear classification head &
7.62M (7,623,802) \\
& & \footnotesize backbone: 7,622,776 \quad head: 1,026 \\

ELM (Entity Linking Model, Phase 2) &
Shared ST-GCN backbone + memory module + linking MLP &
8.34M (8,336,617) \\
& & \footnotesize backbone: 7,622,776 \quad memory: 513,903 \quad linking: 198,912 \quad head: 1,026 \\

CSLR2 backbone (frozen, Phase 3) &
Full CSLR2 architecture &
412.37M (412,371,449) \\
\hline

\end{tabular}

\caption{Parameter counts for all system components. IPN and ELM share a ST-GCN backbone; CSLR2 is kept frozen during integration.}
\label{tab:model_parameters}
\end{table*}

\begin{table*}[!h]
\centering
\small
\setlength{\tabcolsep}{6pt}
\begin{tabular}{p{2.2cm} p{10.5cm}}
\hline
\textbf{Phase} & \textbf{Compute usage} \\
\hline

IPN &
35 runs (Optuna search + 5-seed final sweep), with runtimes ranging from 4--32 minutes per run.
Total compute: $\sim$4.3 GPU-hours. Final model (5 seeds): $\sim$0.7 GPU-hours. \\

ELM &
81 runs (Optuna search + ablation table configurations), with runtimes from 4 minutes to 3.5 hours.
Total compute: $\sim$27.9 GPU-hours. Final reported configurations (6 settings $\times$ $\sim$40 min): $\sim$4 GPU-hours. \\

CSLR2 inference &
Frozen backbone used only for evaluation (no fine-tuning). Evaluation over 55 BOBSL test episodes completes in under 5 minutes. \\

\hline
\end{tabular}
\caption{Compute budget for IPN training, ELM training, and CSLR2 inference. We report the number of runs, runtime ranges, and total GPU-hours per phase. CSLR2 is used in a frozen setting with no fine-tuning.}
\label{tab34}
\end{table*}

\section{Model Size and Budget}
\paragraph{Model Parameters}
\label{app:sizebudget}

IPN and ELM share the same frozen SLGCN backbone weights; ELM adds approximately 714K trainable parameters on top.

\paragraph{Compute Budget}

All experiments were conducted on a single NVIDIA RTX 5090 GPU (24GB VRAM, Blackwell architecture; see Table 15.

\noindent
\textbf{Total reported compute:} $\sim$32 GPU-hours on a single RTX 5090.

\section{Disclaimer: Use of AI-Assisted Development}
\label{app:disclaimer}
Parts of the implementation were developed with assistance from Claude Code (Anthropic), an AI-based coding assistant. AI assistance was used during the development and integration of the IPN and ELM inference pipeline with the CSLR2 baseline. All experimental design, modelling decisions, and scientific conclusions are the authors own.

\section{License and code} 
\label{app:license}
We will release the code upon publication under a standard research license. The use of existing datasets and models in this work is consistent with their intended use for academic research as specified in their original licenses. We do not repurpose any artifacts beyond their permitted research scope.

\end{document}